\pdfoutput=1

\documentclass[11pt]{article}

\usepackage[preprint]{acl}

\usepackage{times}
\usepackage{latexsym}

\usepackage[T1]{fontenc}

\usepackage[utf8]{inputenc}

\usepackage{microtype}

\usepackage{inconsolata}

\usepackage{graphicx}

\usepackage{amsmath,amsfonts,bm}

\def\eqref#1{equation~\ref{#1}}

\def\1{\bm{1}}

\DeclareMathAlphabet{\mathsfit}{\encodingdefault}{\sfdefault}{m}{sl}
\SetMathAlphabet{\mathsfit}{bold}{\encodingdefault}{\sfdefault}{bx}{n}

\newcommand{\reshape}{\textsc{res}}
\newcommand{\df}{\mathcal{D}_\textrm{p}}
\newcommand{\dt}{\mathcal{D}_\textrm{t}}

\newcommand{\llama}{\small Llama3}
\newcommand{\llamaone}{\small Llama3.1}

\newcommand{\replug}{\small \textsc{RePlug-Llama3}}
\newcommand{\replugone}{\small \textsc{RePlug-Llama3.1}}
\newcommand{\PAMQA}{\small \textsc{Pam Qa}}
\newcommand{\kvllamaT}{\small \textsc{KV-Llama3 }}
\newcommand{\kvllamaoneT}{\small \textsc{KV-Llama3.1 }}

\usepackage[utf8]{inputenc}
\usepackage{hyperref}
\usepackage{url}
\usepackage{graphicx}
\usepackage{tabularx}  
\usepackage{booktabs}  
\usepackage{xcolor}    
\usepackage{multirow}  
\usepackage{arydshln}  
\usepackage{subcaption} 
\usepackage{float}
\usepackage{tcolorbox}
\usepackage{wasysym}
\usepackage{amsmath}
\usepackage{amssymb}
\usepackage{algorithm}
\usepackage{algpseudocode}
\usepackage{float}
\usepackage{fancyvrb}

\newtcolorbox{mybox}[2][]{
    colback=white,
    colframe=green!45,
    fonttitle=\bfseries,
    coltitle=black,
    sharp corners,
    title=#2,
    #1
}

\title{Parallel Key-Value Cache Fusion for Position Invariant RAG}

\author{
    {Philhoon Oh} \quad {Jinwoo Shin} \quad {James Thorne}
    \vspace{0.15in}
    \\
    \normalsize\textsc{KAIST AI}
    \vspace{0.1in}
    \\
    {\url{{philhoonoh, jinwoos, thorne}@kaist.ac.kr}} \\
}

\begin{document}
\maketitle
\begin{abstract}


Recent advancements in Large Language Models (LLMs) underscore the necessity of Retrieval Augmented Generation (RAG) to leverage external information. However, LLMs are sensitive to the position of relevant information within contexts and tend to generate incorrect responses when such information is placed in the middle, known as the `Lost in the Middle' phenomenon. In this paper, we introduce a framework that generates consistent outputs for decoder-only models, irrespective of the input context order. Experimental results for three open domain question answering tasks demonstrate position invariance, where the model is not sensitive to input context order and superior robustness to irrelevant passages compared to prevailing approaches for RAG pipelines.

\end{abstract}

\section{Introduction}
In Retrieval Augmented Generation (RAG)~\citep{guu2020realm, lewis2021retrievalaugmented, izacard2022atlasfewshotlearningretrieval}, models first extract relevant information from a knowledge base and then incorporate this extracted information with its parametric knowledge to generate the response. This two-step approach is the de-facto approach for knowledge-intensive tasks~\citep{lewis2021retrievalaugmented, petroni-etal-2021-kilt}.


However, decoder-only models exhibit an intrinsic positional bias, assigning more attention to tokens at the beginning or end of the input sequence while often overlooking relevant context located in the middle, a problem known as the `Lost in the Middle' \citep{Liu2023LostIT}. Previous works to address this issue involve training with specific prompt \citep{he-etal-2024-never} or data-intensive training \citep{an2024makellmfullyutilize}. Other works aimed at modifying positional embeddings \citep{hsieh-etal-2024-found} or reducing positional attention bias in LLMs \citep{yu2024mitigatepositionbiaslarge}. Yet, none of these methods fully guarantee a solution to this intrinsic bias in LLMs for RAG.

\begin{figure}[!t]
  \centering
  \includegraphics[width=1\linewidth]{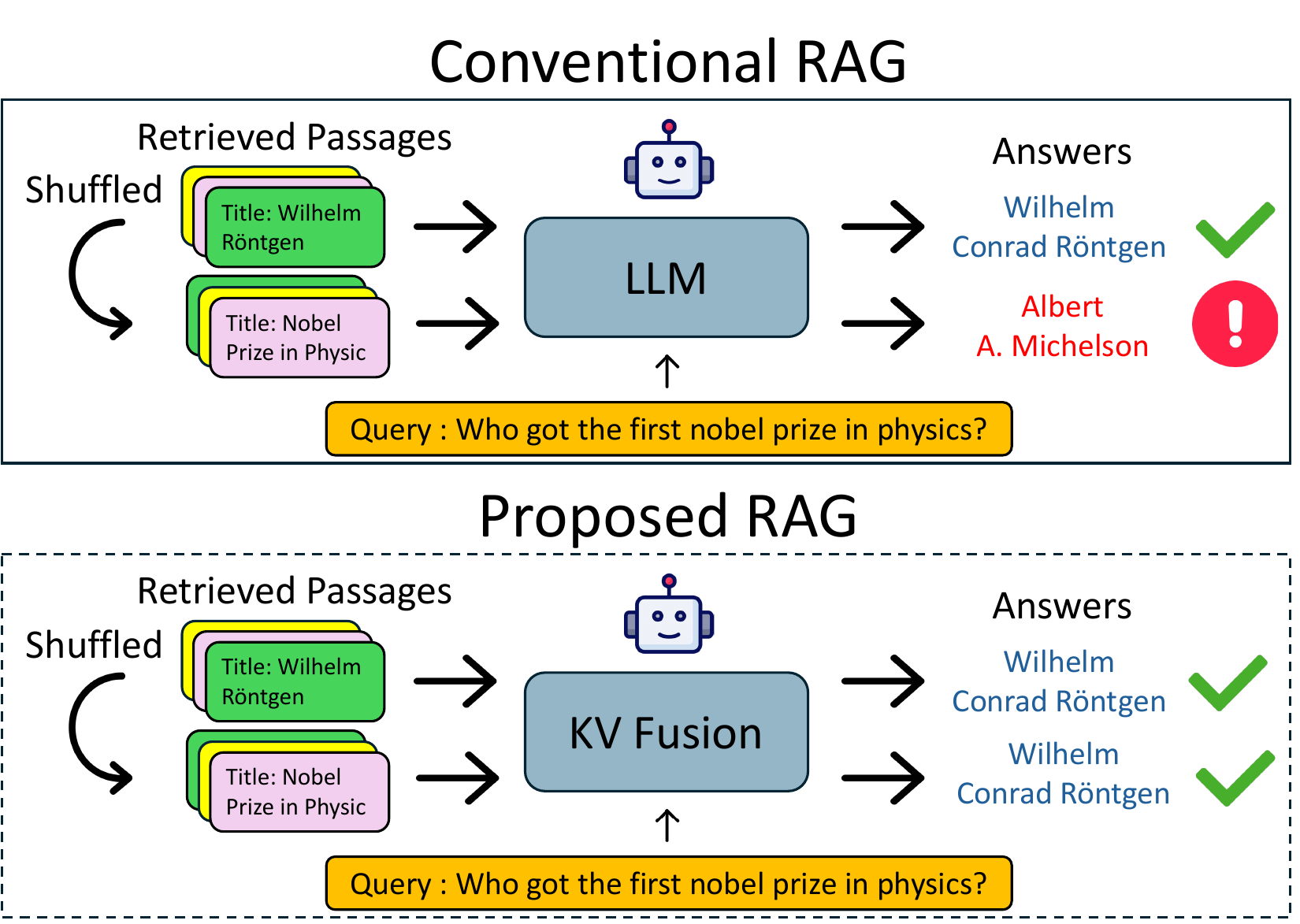}
  \caption{Illustration of the KV-Fusion model: Generated tokens remain consistent even when the retrieved passages are shuffled.}
  \vspace{-1.5em}
  \label{fig:intro}
\end{figure}

\begin{figure*}[t]
    \centering
    \includegraphics[width=0.7\linewidth]{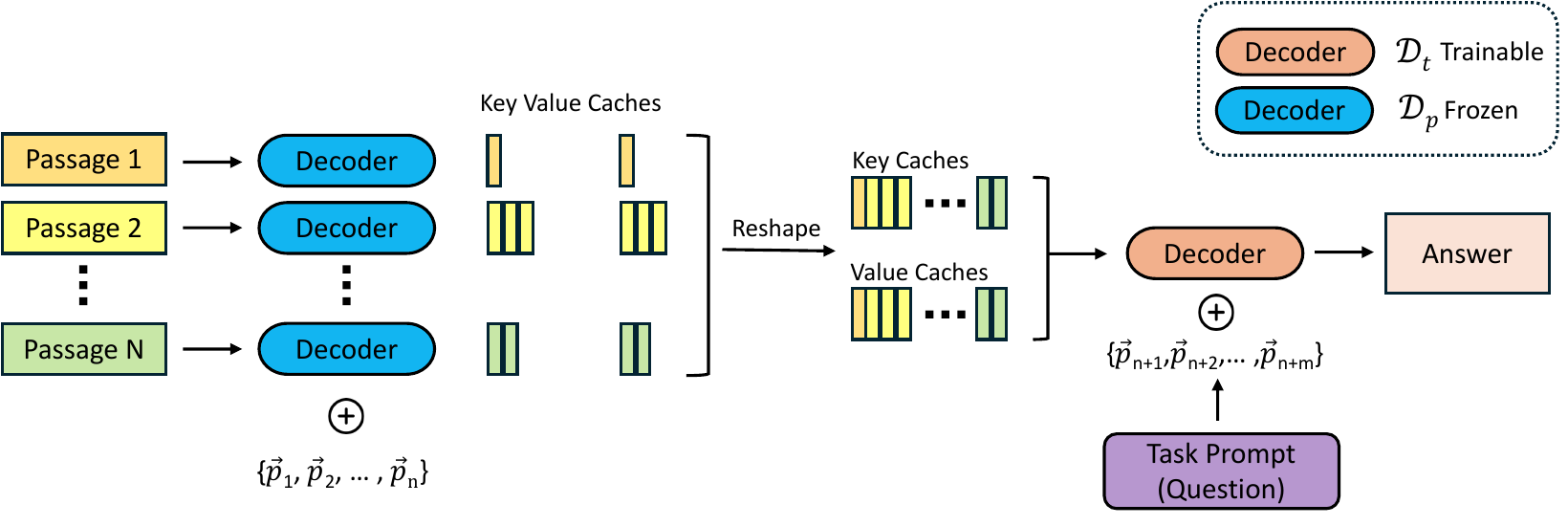}
    \caption{Overview of KV-Fusion Architecture. $\df$ denotes Prefill decoder and $\dt$ represents Trainable decoder. We employ the off-the-shelf LLM to extract the key and value states of the retrieved contexts independently. Then reshaping these caches to train the LLM with task instructions along with questions to generate answers.}
    \label{fig:arch}
    \vspace{-1.5em}
\end{figure*}

In this paper, we introduce a framework for decoder-only models, called \textbf{K}ey \textbf{V}alue \textbf{Fusion} (\textbf{KV Fusion}), to generate consistent outcomes regardless of input order as illustrated in Figure~\ref{fig:intro}. \textbf{KV Fusion} consists of two components: a \textit{prefill} decoder that extracts key-value caches in parallel and a \textit{trainable} decoder that utilizes extracted key-value caches to produce consistent outcomes. This architecture injects uniform positional information into each input context, ensuring consistent output generation even when the input order varies. Experiments on open domain question answering datasets, including NQ \citep{kwiatkowski-etal-2019-natural}, TriviaQA \citep{joshi-etal-2017-triviaqa}, and POPQA \citep{mallen2023trust}, demonstrate KV-Fusion's position-invariant nature, achieving accuracy improvements of 21.4\%, 6.4\%, and 6.6\% over baseline models in shuffled settings. Furthermore, KV-Fusion models exhibit robust and stable accuracies even with additional contexts compared to other approaches.

\label{sec:intro}

\section{Method}



\paragraph{Notation} Our KV-Fusion architecture is illustrated in Figure~\ref{fig:arch}. For clarity, we refer to this prefill decoder as $\df$, characterized by the number of key and value heads $|\mathit{H}|$, each with a $\mathit{d}_h$ dimension. We denote the trainable decoder as $\dt$, and represent the set of input passages as $\mathcal{C} = \{c_1, c_2, \dots, c_N\}$ with fixed token length $n$ for each $c_i$. This set of passages represents smaller chunks of a long document or retrieved contexts. Lastly, let $L$ represent the total number of layers in $\df$ and $\dt$, and let $l$ denote the $l$th layer.



\paragraph{Prefill Decoder ($\df$)} extracts the KV-caches from multiple input passages in parallel, resulting in the injection of identical local positional embeddings ${\{\vec{p_1}, \vec{p_2}, \dots, \vec{p_n}\}}$. The layer-wise cache representation for each $c_i$ is as follows:
\begin{equation*}
\{k^{l}_{i}, v^{l}_{i}\}_{l=1}^L = \df(c_i), \quad k^{l}_{i}, v^{l}_{i} \in \mathbb{R}^{|\mathit{H}| \times n \times \mathit{d}_h}
\end{equation*}
Next, we reshape layer-wise KV-caches by concatenating along the token axis over $N$ contexts, forming a single cache for each layer $l$:
\begin{equation*}
K^{l} = \reshape(\{k^{l}_{i}\}_{i=1}^N) \quad
V^{l} = \reshape(\{v^{l}_{i}\}_{i=1}^N)
\end{equation*}
Here, $K^{l}, V^{l} \in \mathbb{R}^{|\mathit{H}| \times (N \times n) \times \mathit{d}_h}$ are reshaped KV-cache for the corresponding layer over input passages. These caches prefill and serve as grounding knowledge for training $\dt$.



\paragraph{Trainable Decoder ($\dt$)} takes two inputs: (1) reshaped KV-caches ($\{K^{l}, V^{l}\}_{l=1}^L$) and (2) target tokens, which contain instruction queries, and answers with a length of $m$ tokens. To ensure sequential alignment of positional information with the KV-caches, position information starting from $\vec{p}_{n+1}$ to $\vec{p}_{n+m}$ are assigned. We then train $\dt$ using next-token prediction, conditioning on the reshaped KV-caches rather than previous tokens: 
\begin{equation*} 
\dt(y| q, \mathcal{C^\prime}) \triangleq \dt(y| q, \{K^{l}, V^{l}\}_{l=1}^L)
\end{equation*} 
Here, $q$ denotes the instruction with query tokens, and $y$ is answer tokens. $\mathcal{C^\prime}$ represents the set of input passages tokens, and $\{K^{l}, V^{l}\}_{l=1}^L$ is the reshaped KV-caches corresponding to $\mathcal{C^\prime}$. We illustrate the details of KV-Fusion in Appendix~\ref{appendix:pseudo}

\label{sec:method}

\section{Experiment Setup}
\begin{figure*}[!ht]
    \centering
    \begin{subfigure}{0.3\textwidth}
        \centering
        \caption{\scriptsize NQ - Llama3}
        \includegraphics[width=\textwidth]{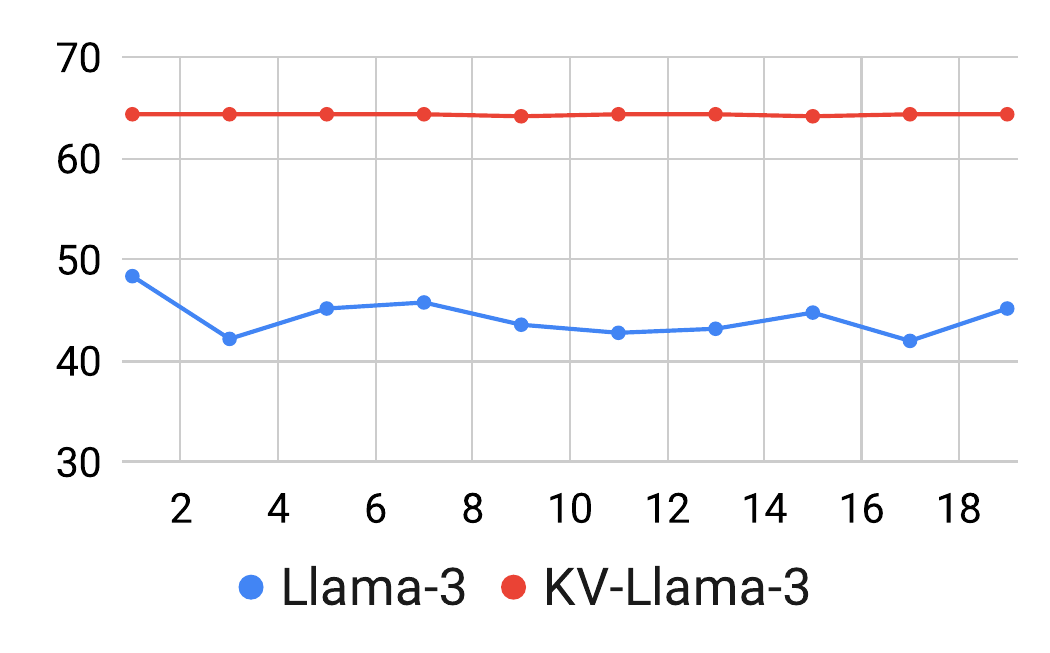}
    \end{subfigure}
    \hfill
    \begin{subfigure}{0.3\textwidth}
        \centering
        \caption{\scriptsize TQA - Llama3}
        \includegraphics[width=\textwidth]{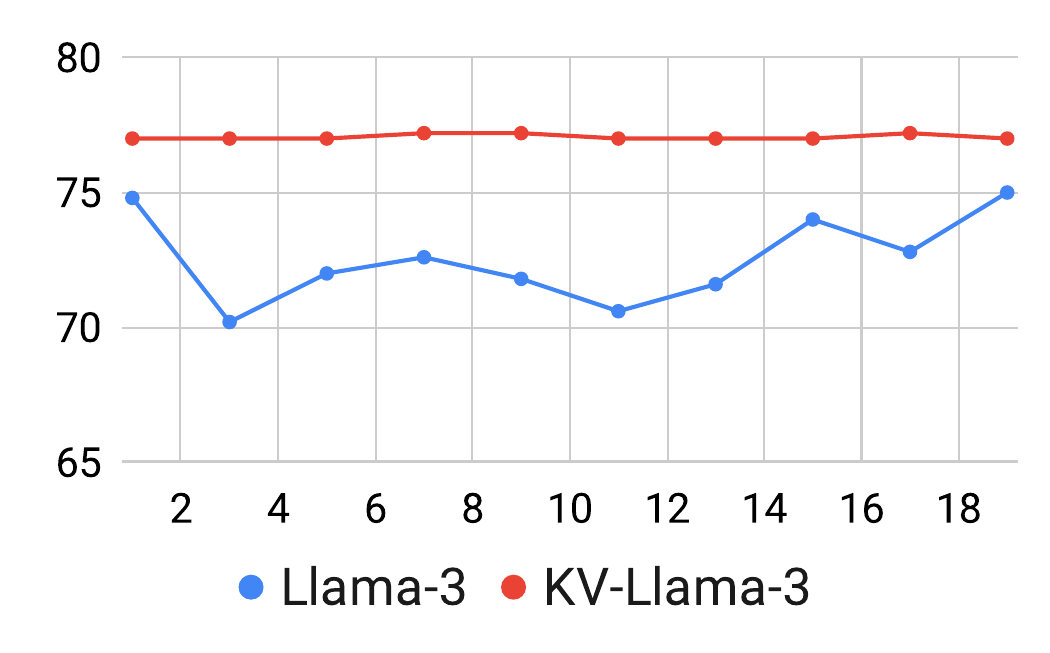}
    \end{subfigure}
    \hfill
    \begin{subfigure}{0.3\textwidth}
        \centering
        \caption{\scriptsize POPQA - Llama3}
        \includegraphics[width=\textwidth]{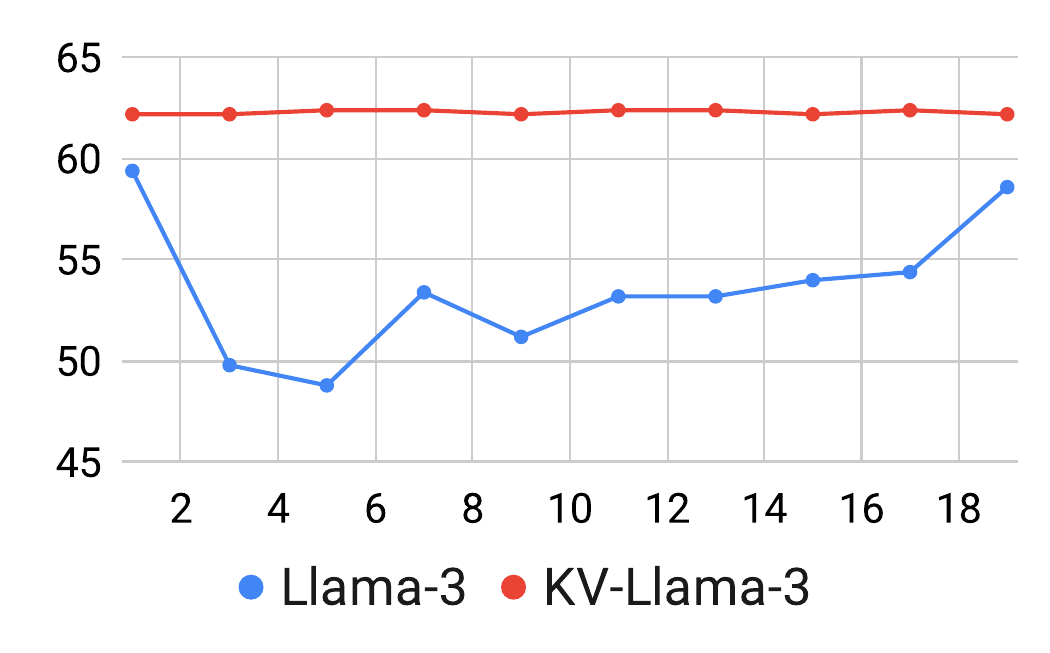}
    \end{subfigure}



    \caption{Comparison of EM Accuracy between \textcolor{red}{KV-Llama3} and \textcolor{blue}{Llama3} across different gold context positions. KV-Llama3 maintains its accuracy, while Llama3 shows a tendency for the `lost in the middle' problem.}
    \vspace{-1em}
    \label{fig:exp1-1}
\end{figure*}


\begin{figure*}[!ht]
    \centering
    \begin{subfigure}{0.3\textwidth}
        \centering
        \caption{\scriptsize NQ}
        \includegraphics[width=\textwidth]{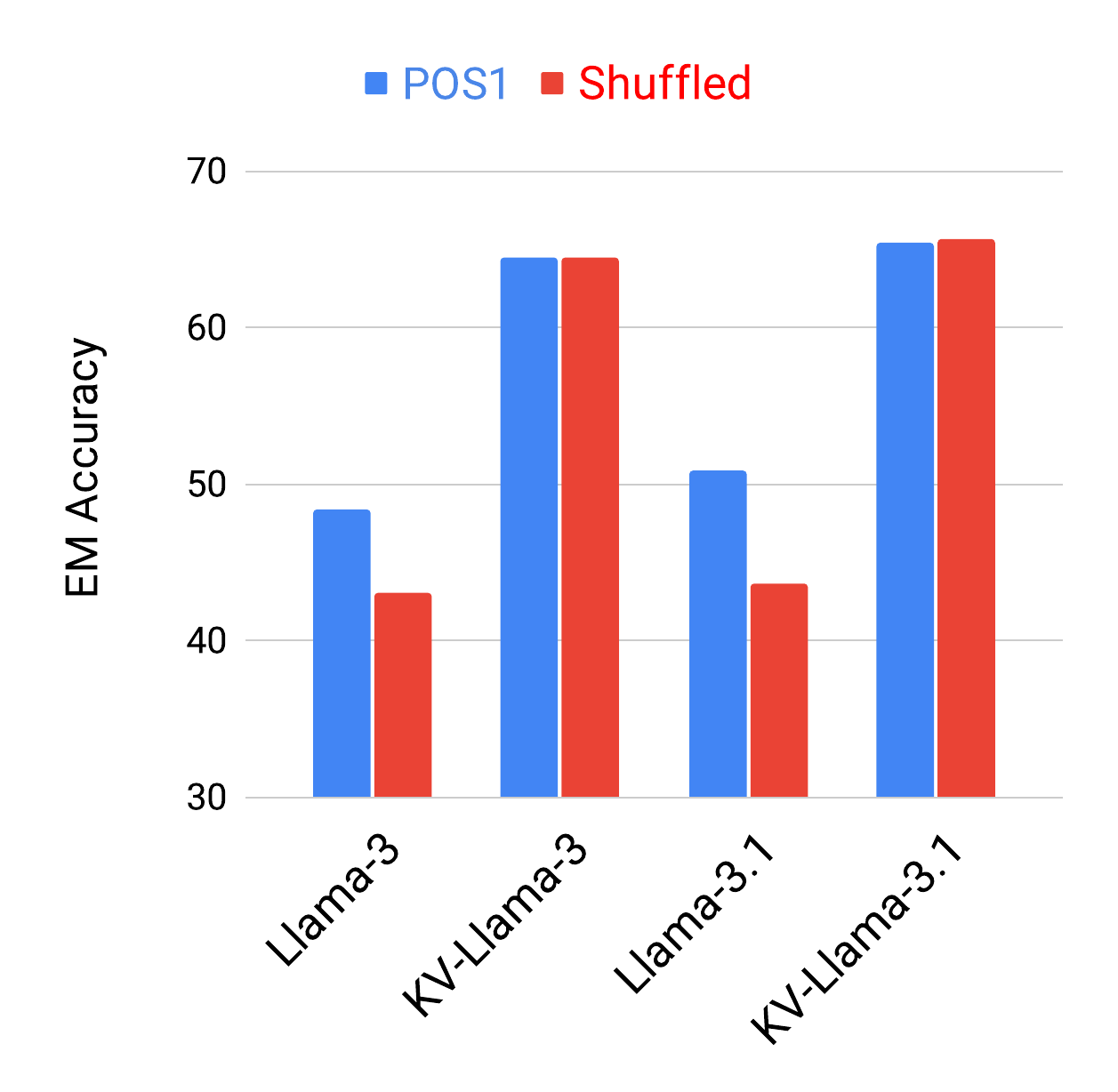}
    \end{subfigure}
    \hfill
    \begin{subfigure}{0.3\textwidth}
        \centering
        \caption{\scriptsize TQA}
        \includegraphics[width=\textwidth]{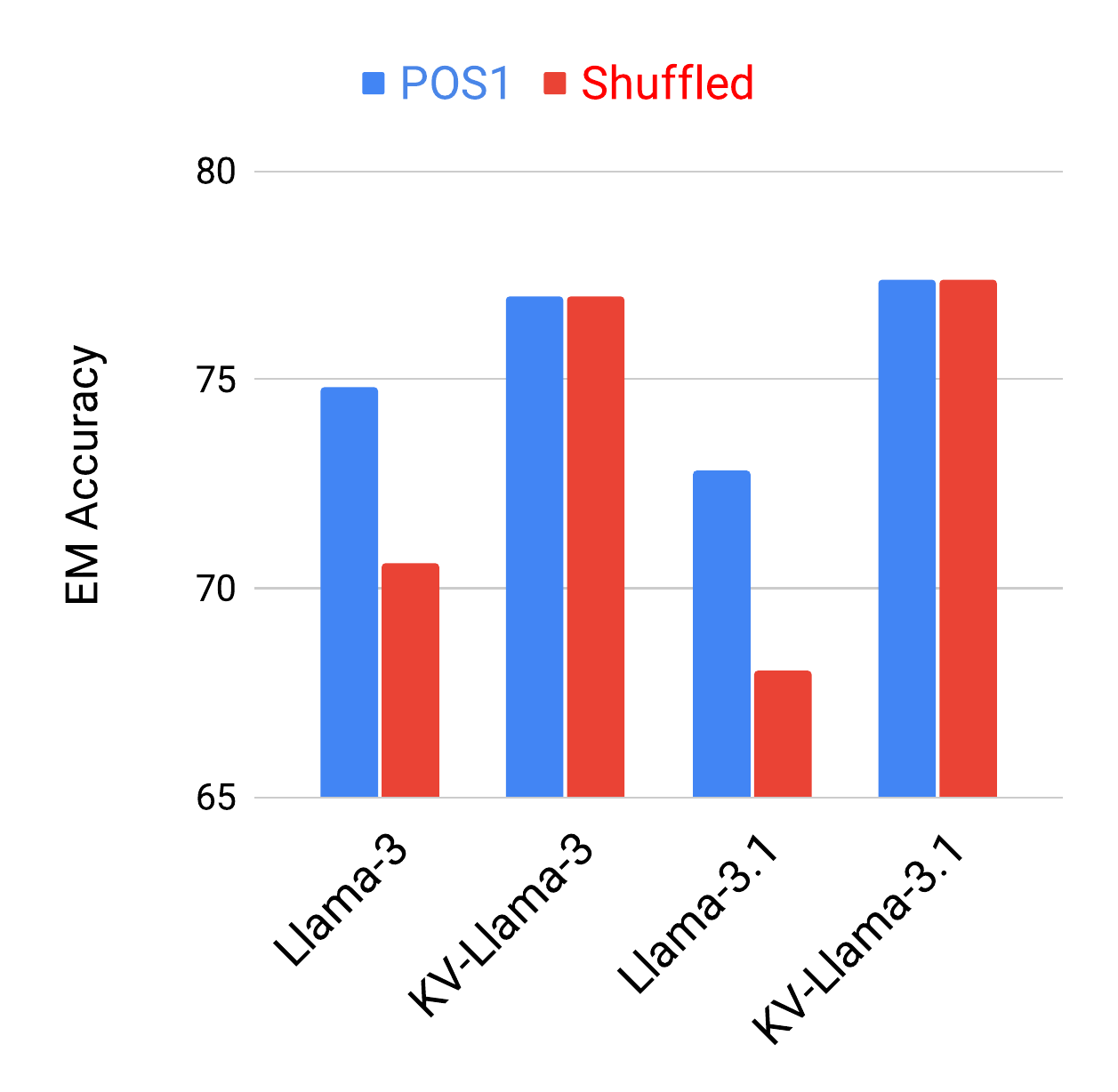}
    \end{subfigure}
    \hfill
    \begin{subfigure}{0.3\textwidth}
        \centering
        \caption{\scriptsize POPQA}
        \includegraphics[width=\textwidth]{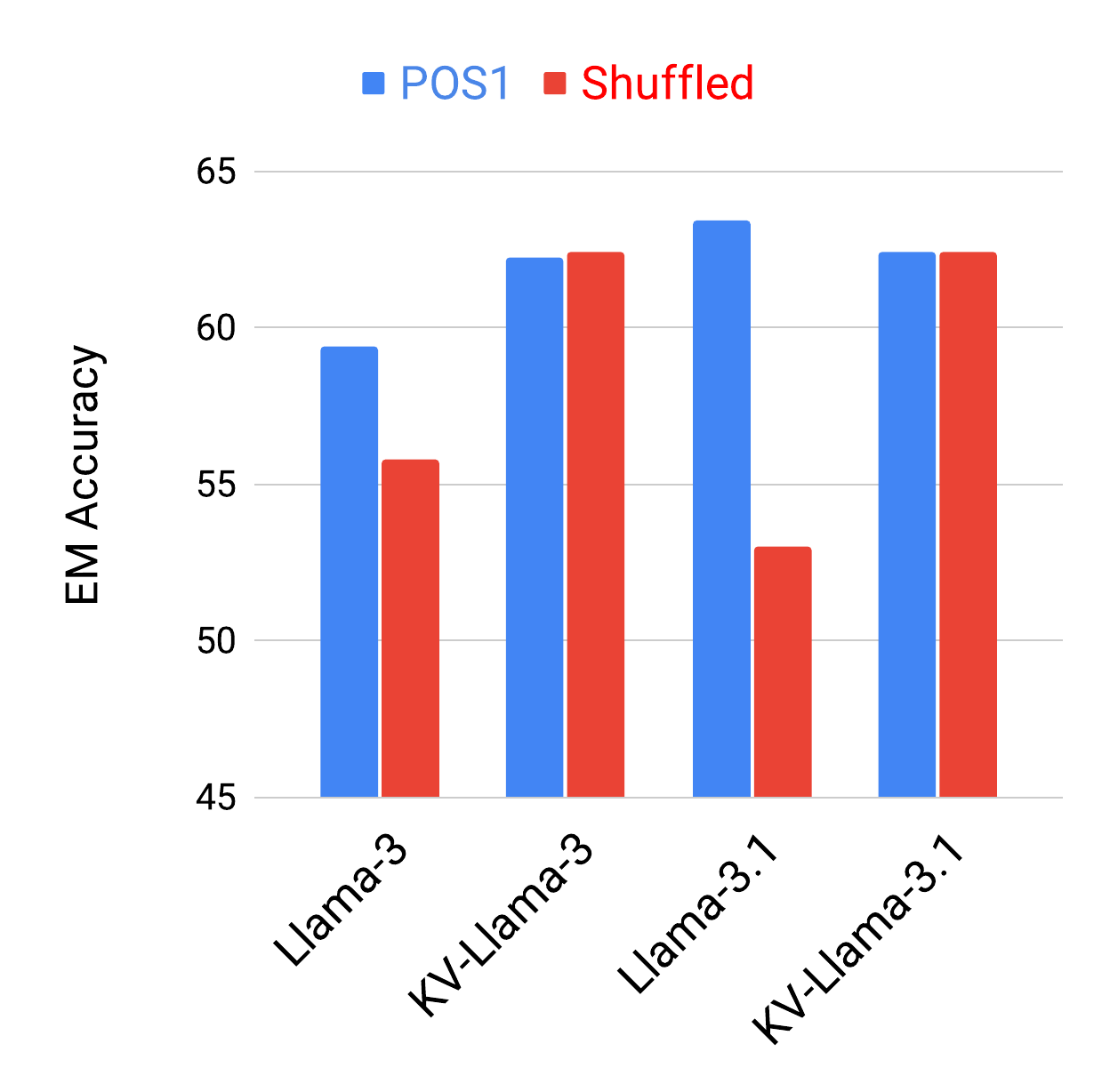}
    \end{subfigure}
    \caption{Accuracies of baseline and KV models in two scenarios: 1) \textcolor{blue}{POS1}, where the gold context is positioned first, and 2) \textcolor{red}{Shuffled}, where contexts are randomly ordered. KV models maintain their accuracy on both cases, while baseline models struggle in shuffled setting, leading in a wider accuracy gap between the baseline and KV-models.}
    \label{fig:exp1-2}
    \vspace{-1em}
\end{figure*}

\subsection{Datasets}\label{experiments:dataset}
 We consider three open domain question answering datasets: Natural Questions \citep{kwiatkowski-etal-2019-natural}, TriviaQA \citep{joshi-etal-2017-triviaqa}, and POPQA \citep{mallen2023trust} .\footnote{Note that NQ and TriviaQA are filtered version from DPR~\citep{karpukhin-etal-2020-dense}} For the base retrieval corpus, we utilize a December 2018 Wikipedia snapshot consisting of 21 million passages, following \citep{yen2024longcontextlanguagemodelingparallel, yu2024rankragunifyingcontextranking}. Lastly, we use the DPR \citep{karpukhin-etal-2020-dense} as our baseline retriever to extract the top 40 passages for each dataset. \footnote{Note that we used DPR trained from scratch with its hard negatives on the December 2018 Wikipedia snapshot}

\paragraph{Dataset Construction}\label{experiments:dataset2}
To enhance the robustness in RAG, we train models with irrelevant contexts \citep{fang-etal-2024-enhancing, yoran2024making}. To this end, we draw the best gold context and extract key phrases among candidate passages by prompting GPT-4o API with a fine-grained template. If all responses are negative, the instance is discarded. Otherwise, we retain the extracted key phrases as evidence, which is later used for training. Negative contexts are sampled from DPR-retrieved passages that do not contain any answer. Each training instance consists of one gold context and 19 negative contexts. The prompt for this process and statistics of all datasets are described in Appendix~\ref{appendix:data_cons}.



\paragraph{Metric and Evaluations}
Exact Match (EM) Accuracy is used for evaluation. \citep{asai2023selfraglearningretrievegenerate, mallen2023trust}. However, we observe that as more documents are added to the input, baseline models tend to generate intrinsic knowledge or hallucinated responses~\cite{hsieh2024ruler}. To address this, we incorporate answerability into the prompt, requiring responses to be concise and limited to a single sentence. Lastly, we set a 48-token limit and use greedy-decoding~\cite{huang2023surveyhallucinationlargelanguage} for baselines and KV-Fusions. The template and example for baseline are illustrated in Appendix~\ref{appendix:baseline_format}.

\subsection{Training}
\paragraph{Input Formatting}
Each input passage is formatted with `{\small\texttt{Title:}}\{title\}' and `{\small\texttt{Context:}}\{text\}', followed by a document boundary, {\small\texttt{`===='}}. For target tokens, we preprend a signal token, {\small{\texttt{$<$|question\_answering|$>$}}}, to guide the model's behavior during inference \citep{asai2023selfraglearningretrievegenerate}. Next, we append instruction and `{\small\texttt{Question:}}\{question\}'. Finally, we add answer tokens, which contain both an answer string and a key phrase as evidence, as described in Section~\ref{experiments:dataset2}. We hypothesize that appending key phrases enhances the model's robustness~\citep{thoppilan2022lamdalanguagemodelsdialog, menick2022teachinglanguagemodelssupport}. Format examples are provided in Appendix~\ref{appendix:format}.

\paragraph{Technical Details}
We initialize both $\df$ and $\dt$ with the Llama3-8B model~\citep{LLAMA3}. We fine-tune each dataset with a maximum learning rate of $2 \times 10^{-5}$ using the AdamW. Across all datasets, we use a batch size of 64 on four A100(80G) GPUs. For the NQ and TQA datasets, models are trained for two epochs. For the POPQA dataset, we fine-tune it on top of the TQA fine-tuned model due to its small training size. The same procedure is applied to the Llama3.1-8B. Detail hyperparameters are reported in Appendix~\ref{appendix:hyper}.

\label{sec:exp}

\section{Results}
\begin{table*}[!t]
    \renewcommand{\arraystretch}{1.15}
    \centering
    \small
    \begin{tabularx}{\textwidth}{p{3.1cm} XXXXX XXXXX XXXXX}
        \toprule
        \textbf{Dataset}    & \multicolumn{4}{c}{\textbf{NQ}}   & \multicolumn{4}{c}{\textbf{TQA}}  & \multicolumn{4}{c}{\textbf{POPQA}}  \\
        \cmidrule(lr){2-5} \cmidrule(lr){6-9} \cmidrule(lr){10-13} 
        \textbf{Top-K} & 5 & 10 & 20 & 40 & 5 & 10 & 20 & 40 & 5 & 10 & 20 & 40    \\
        \midrule
        \llama  &\scriptsize{34.1}  &\scriptsize{37.2} &\scriptsize{40.4} &\scriptsize{38.8}  &\scriptsize{62.4} &\scriptsize{66.5} &\scriptsize{67.2} &\scriptsize{64.7}  &\scriptsize{31.7} &\scriptsize{33.7} &\scriptsize{33.7} &\scriptsize{31.6}\\
        \llamaone  &\scriptsize{43.2}  &\scriptsize{41.5} &\scriptsize{42.7} &\scriptsize{42.4}   &\scriptsize{64.0} &\scriptsize{64.8} &\scriptsize{65.9} &\scriptsize{67.1} &\scriptsize{31.1} &\scriptsize{33.4} &\scriptsize{32.8} &\scriptsize{33.3}\\
        \hdashline
        \replug &\scriptsize{35.6}  &\scriptsize{34.0} &\scriptsize{33.6} &\scriptsize{32.2}  &\scriptsize{57.7}  &\scriptsize{56.4} &\scriptsize{55.8} &\scriptsize{56.3} &\scriptsize{30.1}  &\scriptsize{28.1} &\scriptsize{26.0} &\scriptsize{26.7} \\
        \replugone  &\scriptsize{38.6}  &\scriptsize{36.4} &\scriptsize{35.3} &\scriptsize{34.1}  &\scriptsize{65.1}  &\scriptsize{64.4} &\scriptsize{62.8} &\scriptsize{60.6} &\scriptsize{35.4}  &\scriptsize{33.8} &\scriptsize{30.7} &\scriptsize{29.4} \\
        \PAMQA  &\textbf{\scriptsize{51.9}}  &\scriptsize{46.5} &\scriptsize{40.7} &\scriptsize{19.7}  &\scriptsize{65.9}  &\scriptsize{61.2} &\scriptsize{52.3} &\scriptsize{29.0} &\scriptsize{37.0} &\scriptsize{35.9} &\scriptsize{34.9} &\scriptsize{15.7} \\
        \kvllamaT  &\scriptsize{51.6}  &\scriptsize{51.7} &\textbf{\scriptsize{51.4}} &\textbf{\scriptsize{49.8}} &\scriptsize{67.5}  &\textbf{\scriptsize{68.8}} &\textbf{\scriptsize{69.3}} &\textbf{\scriptsize{69.3}} &\scriptsize{44.5} &\scriptsize{46.7} &\textbf{\scriptsize{48.3}} &\textbf{\scriptsize{46.7}} \\
        \kvllamaoneT  &\scriptsize{51.7}  &\textbf{\scriptsize{51.8}} &\scriptsize{50.8} &\scriptsize{49.0}  &\textbf{\scriptsize{68.6}}  &\scriptsize{68.3} &\scriptsize{69.3} &\scriptsize{68.7} &\textbf{\scriptsize{44.7}} &\textbf{\scriptsize{47.6}} &\scriptsize{47.4} &\scriptsize{45.3} \\
        \bottomrule
    \end{tabularx}
    \caption{Percentage accuracy comparison between position-invariant readers. Across top-k results, KV-fusion maintains stable and the strong accuracies, while other models either degrade or exhibit relatively low accuracies.}
    \vspace{-5mm}
    \label{tab:exp2}
\end{table*}

\begin{table}[!t]
\centering
\begin{tabular}{@{}lccc@{}}
\toprule
{\small \textbf{Model}}       & {\small \textbf{NQ}} & {\small \textbf{TQA}} & {\small \textbf{POPQA}} \\ \midrule
\llama              & {\small 17.6}        & {\small 14.4}        & {\small 8.2}           \\
\llamaone            & {\small 14.6}        & {\small 10.6}        & {\small 4.6}           \\
\kvllamaT           & {\small 99.6}        & {\small 99.2}        & {\small 99.4}          \\
\kvllamaoneT         & {\small 99.6}        & {\small 99.8}        & {\small 99.6}          \\ \bottomrule
\end{tabular}
\caption{Percentage Token-Level Match of KV-Fusion and baseline models. While baselines produce varying tokens when contexts are shuffled, KV-Fusion generates identical tokens, even with order perturbations.}
\vspace{-1.5em}
\label{fig:ab1}
\end{table}



\paragraph{Position Invariant RAG}\label{sub:pos}

To demonstrate the position-agnostic property, we test models with the gold context placed at varying positions. For each dev dataset, we construct 10 versions by inserting gold context at every alternate location (1st, 3rd, etc.), along with an additional dev set where all 20 contexts are randomly shuffled. To manage the increased inference time, we evaluate the first 500 instances. As shown in Figure~\ref{fig:exp1-1}, KV-Llama3 maintains consistent accuracy across all datasets, regardless of the position of the gold context, while conventional Llama3 shows varying accuracy. A similar pattern is observed with KV-Llama3.1 and Llama3.1 as shown in Appendix~\ref{appendix:llama3.1}. Figure~\ref{fig:exp1-2} emphasizes this difference: the accuracy of the baseline model drops considerably with shuffled contexts, while the KV models maintain stable performance. In the shuffled scenario, KV-Llama3 achieves higher accuracy than baselines on the NQ, TQA, and POPQA datasets, with similar trends observed for KV-Llama3.1. These findings suggest that KV-Fusion improves performance in RAG.

\paragraph{Comparison with Recent Methods}\label{sub:comp}

We evaluate KV models alongside other position-agnostic methods: PAM QA \citep{he-etal-2024-never}, which employs multi-step reasoning to reduce position bias, and \textsc{RePlug} \citep{shi-etal-2024-replug}, which predicts the next token based on a weighted score for each context. Across the test sets, we utilize up to 40 DPR-retrieved passages, using default settings for PAM QA and the same configurations as Llama3 for \textsc{RePlug}. As shown in Table~\ref{tab:exp2}, KV-models achieve the highest accuracy across datasets except the top-5 NQ case. Notably, KV models, originally trained with 20 passages, demonstrate strong robustness even with the top 40 passages. PAM QA performs well with up to 20 passages but shows an average accuracy decline of 50.3\% when scaled to the top 40. \textsc{RePlug} follows a similar pattern as the baselines but also experiences performance degradation. Comparable results are observed with contriever passages as shown in Appendix~\ref{appendix:contriever}. These results indicate that KV-Fusion enhances robustness even with large input passages within the RAG pipeline.

\label{sec:res}

\section{Ablations}
\paragraph{Token-Level Consistency}\label{sub:token-level} Previous research \citep{he-etal-2024-never, hsieh-etal-2024-found, yu2024mitigatepositionbiaslarge} on positional bias study the effect of performance (e.g. accuracy) under perturbation of the input.  Models are regarded as consistent if performance is stable regardless of the position of the gold context. In this section, we further investigate target token-level consistency. We define Token-Level Match (TLM) as the accuracy of Exact Match (EM) between responses in two scenarios: (1) POS1, where the gold context is at the top, and (2) Shuffled, with randomly ordered contexts. Here, $p$ and $s$ denote the responses in POS1 and Shuffled settings, and $N$ is the total number of instances. We experiment with the top-20 dev set in Section~\ref{sec:res}.

{\small
\[
    TLM(p, s, N) = \frac{1}{N}{\sum_{i=1}^{N} EM(p_i = s_i)} 
\]
}


The results are shown in Table~\ref{fig:ab1}. Using greedy-decoding (as this is non-stochastic), Llama3 and Llama3.1 are sensitive to positional context, resulting in low TLM scores across datasets. Notably, the models are most sensitive to positional bias on POPQA. 
In contrast, KV-Fusion demonstrates nearly perfect TLM scores across all datasets (all $>99\%$). This indicates that KV-Fusion generates near identical tokens regardless of context-order perturbations, highlighting position-agnostic behavior and maintaining token-level consistency where context order is perturbed.

\paragraph{Comparison with Rerankers}\label{sub:token-level} To enhance the end-to-end performance of the RAG pipeline, rerankers \cite{nogueira-etal-2020-document, 10.1145/3539618.3592047, 10.1145/3626772.3657951} are often used to prioritize relevant context while filtering out irrelevant ones \citep{glass-etal-2022-re2g, yu2024rankragunifyingcontextranking}. However, we hypothesize that a robust model capable of handling irrelevant passages could eliminate the need for a reranking module. To test this, we rerank the top 40 retrieved passages from the test sets in Section~\ref{sec:res} and retain only the top 20 passages, discarding the lower-ranked ones. We utilize point-wise rerankers: MonoT5-3B\cite{nogueira-etal-2020-document}, Rankt5-3B\cite{10.1145/3539618.3592047}, and RankLLama-7B\cite{10.1145/3626772.3657951}. We then compared the EM accuracy of baseline models evaluated on these top-20 reranked test sets with KV-Fusion models evaluated on the original top-40 test sets. 

Results are described in Table~\ref{fig:ab2}. Ranking and truncation improve the accuracy of Llama3 across datasets. However, LLama3.1 exhibits slight gains in POPQA but a minor decline in NQ, TQA. This behavior is likely due to LLama3.1's ability to process longer sequences, making reranking and truncation less beneficial. In contrast, KV-Fusion demonstrates consistently superior accuracy across all datasets, showing its robustness and suggesting that reranking may not be necessary for the RAG.

\label{sec:ablations}

\section{Related Works}
\begin{table}[!t]
\centering
\begin{tabular}{@{}lccc@{}}
\toprule
{\small \textbf{Model}}       & {\small \textbf{NQ}} & {\small \textbf{TQA}} & {\small \textbf{POPQA}} \\ \midrule
\llama {\small$\ast$}              & {\small 38.8}        & {\small 64.7}        & {\small 31.6}           \\ \hdashline
\small{+ monot5}            & {\small 40.8}        & {\small 67.0}       & {\small 35.0}          \\
\small{+ rankt5}            & {\small 41.4}       & {\small 67.2}       & {\small 35.4}          \\
\small{+ rankllama}         & {\small 40.6}       & {\small 67.4}       & {\small 35.2}           \\ \midrule
\llamaone {\small$\ast$}           & {\small 42.4}        & {\small 67.1}        & {\small 33.3}           \\ \hdashline
\small{+ monot5}            & {\small 41.5}        & {\small 65.5}        & {\small 34.8}          \\
\small{+ rankt5}            & {\small 42.8}       & {\small 65.9}       & {\small 35.6}           \\
\small{+ rankllama}         & {\small 42.0}       & {\small 66.1}       & {\small 35.4}          \\ \midrule
\kvllamaT {\small$\ast$}         & {\small 49.8}       & {\small 69.3}       & {\small 46.7}          \\
\kvllamaoneT {\small$\ast$}        & {\small 49.0}       & {\small 68.7}       & {\small 45.3}          \\ \bottomrule
\end{tabular}
\caption{Accuracy comparison of rerankers based on top-20 reranked contexts, with non-reranked ones using original top-40 contexts ($\ast$). Results show KV-Fusion eliminates the need for ranking in the RAG pipeline.}
\vspace{-1.5em}
\label{fig:ab2}
\end{table}

\paragraph{Retrieval Augmented Generation (RAG)}\label{sub:comp} With recent advancements in LLMs\citep{geminiteam2024gemini15unlockingmultimodal, GPT-4o}, Retrieval Augmented Generation (RAG) have proven to be effective in complementing LLMs across various tasks: managing long-tail information~\citep{mallen2023trust}, reducing hallucinations~\citep{huang2023survey, shi-etal-2024-trusting}, and improving interpretability~\citep{borgeaud2022improvinglanguagemodelsretrieving, rudin2021interpretable}. The idea of utilizing external knowledge has become prevalent, particularly in knowledge-intensive~\citep{thorne-etal-2018-fever, lewis2021retrievalaugmented, petroni-etal-2021-kilt}, where retrievers like DPR and Contriever~\citep{karpukhin-etal-2020-dense, izacard2021contriever} first retrieve relevant information, and readers like FiD, ATLAS~\citep{izacard2020leveraging, izacard2022atlasfewshotlearningretrieval} incorporate the retrieved information to make predictions.

\paragraph{Robustness and Bias in RAG Pipeline}\label{sub:comp} Despite the promising capabilities of the RAG system, one major challenge is the notable drop in performance when irrelevant contexts exist during inference. \citep{shi2023largelanguagemodelseasily, oh-thorne-2023-detrimental}, along with incorrect responses even when the gold context appears in the middle \citep{Liu2023LostIT}. To address these issues, \citealt{xu2023recompimprovingretrievalaugmentedlms} trained an auxiliary LLM to summarize and extract relevant contexts, while \citealt{yoran2024makingretrievalaugmentedlanguagemodels} proposed a simple Natural Language Inference (NLI) model to eliminate unnecessary passages. Also, \citealt{he-etal-2024-never} suggests decomposing inference into multi-step reasoning, enabling the model to generate accurate responses regardless of the context order. Other methods focus on internal features, such as adjusting position hidden states or calibrating attention biases \citep{hsieh-etal-2024-found, yu2024mitigatepositionbiaslarge}. However, none of these approaches fully resolve a complete solution for the `Lost in the Middle' problem.

\paragraph{Key Value Cache in RAG}\label{sub:comp} Recent studies in Key-Value Cache (KV cache) in RAG focus on improving long context understanding and runtime efficiency. For example, RAGCache \cite{jin2024ragcacheefficientknowledgecaching} leverages precomputed KV caches for retrieval and faster inference. Similarly, Cache Augmented Generation (CAG) \cite{chan2024dontragcacheaugmentedgeneration} augments the knowledge via KV caches, demonstrating its efficiency for long-context understanding. TurboRAG \cite{lu2024turboragacceleratingretrievalaugmentedgeneration} utilizes KV caches for training the RAG pipeline, whereas our work focuses on using KV caches to address the `Lost in the Middle' problem, enhancing robustness and enabling an order-invariant pipeline.

\label{sec:related}

\section{Conclusion}
This paper presents KV-Fusion, a lightweight training scheme that addresses positional bias and improves the robustness of decoder-only models in the RAG pipeline. KV-Fusion trains language models to be context-order invariant by extracting and pre-filling KV caches with identical positional information and then training decoder-only models using these caches. The results highlight not only the robustness of KV-Fusion in handling a large number of input passages but also its position-invariant property. Our empirical evaluations on three open-domain datasets indicate that KV-Fusion can improve the performance and reliability of the RAG system.

\label{sec:conclusion}

\section{Limitations}
One limitation of this work is its focus on question answering. Although the most common dataset for evaluating LLMs' understanding with large context would be the needle-in-a-haystack (NIAH) dataset \cite{niah}, our experiments are centered around question-answering, which is more challenging than NIAH \cite{hsieh2024ruler}.

Second limitation is that our experiments are limited to single-hop question answering, where multi-step reasoning is not required. For example, datasets like HotpotQA \cite{yang2018hotpotqa}, and MuSiQue \cite{trivedi-etal-2022-musique} require multiple passages to derive answers. This work, however, focuses on single-hop question-answering datasets, making it difficult to assess the impact of KV-fusion in multi-hop datasets.

Third limitation is that this work does not fully explore the use of KV-cache for training LLMs. Recently, training LLMs by conditioning key-value caches have gained attention \cite{sun2024cacheoncedecoderdecoderarchitectures} for its efficiency, though our approach remains underexplored in terms of language modeling. However, we present strong empirical results to solve the `Lost in the middle' problem. We hope our work can facilitate future studies on utilizing key-value cache for training LLMs.

\label{sec:limitations}

\bibliography{acl_latex}

\begin{thebibliography}{50}
\providecommand{\natexlab}[1]{#1}

\bibitem[{An et~al.(2024)An, Ma, Lin, Zheng, and Lou}]{an2024makellmfullyutilize}
Shengnan An, Zexiong Ma, Zeqi Lin, Nanning Zheng, and Jian-Guang Lou. 2024.
\newblock \href {https://arxiv.org/abs/2404.16811} {Make your llm fully utilize the context}.
\newblock \emph{Preprint}, arXiv:2404.16811.

\bibitem[{Asai et~al.(2023)Asai, Wu, Wang, Sil, and Hajishirzi}]{asai2023selfraglearningretrievegenerate}
Akari Asai, Zeqiu Wu, Yizhong Wang, Avirup Sil, and Hannaneh Hajishirzi. 2023.
\newblock \href {https://arxiv.org/abs/2310.11511} {Self-rag: Learning to retrieve, generate, and critique through self-reflection}.
\newblock \emph{Preprint}, arXiv:2310.11511.

\bibitem[{Borgeaud et~al.(2022)Borgeaud, Mensch, Hoffmann, Cai, Rutherford, Millican, van~den Driessche, Lespiau, Damoc, Clark, de~Las~Casas, Guy, Menick, Ring, Hennigan, Huang, Maggiore, Jones, Cassirer, Brock, Paganini, Irving, Vinyals, Osindero, Simonyan, Rae, Elsen, and Sifre}]{borgeaud2022improvinglanguagemodelsretrieving}
Sebastian Borgeaud, Arthur Mensch, Jordan Hoffmann, Trevor Cai, Eliza Rutherford, Katie Millican, George van~den Driessche, Jean-Baptiste Lespiau, Bogdan Damoc, Aidan Clark, Diego de~Las~Casas, Aurelia Guy, Jacob Menick, Roman Ring, Tom Hennigan, Saffron Huang, Loren Maggiore, Chris Jones, Albin Cassirer, Andy Brock, Michela Paganini, Geoffrey Irving, Oriol Vinyals, Simon Osindero, Karen Simonyan, Jack~W. Rae, Erich Elsen, and Laurent Sifre. 2022.
\newblock \href {https://arxiv.org/abs/2112.04426} {Improving language models by retrieving from trillions of tokens}.
\newblock \emph{Preprint}, arXiv:2112.04426.

\bibitem[{Chan et~al.(2024)Chan, Chen, Cheng, and Huang}]{chan2024dontragcacheaugmentedgeneration}
Brian~J Chan, Chao-Ting Chen, Jui-Hung Cheng, and Hen-Hsen Huang. 2024.
\newblock \href {https://arxiv.org/abs/2412.15605} {Don't do rag: When cache-augmented generation is all you need for knowledge tasks}.
\newblock \emph{Preprint}, arXiv:2412.15605.

\bibitem[{Dubey et~al.(2024)Dubey, others Abhinav~Jauhri, Pandey, Kadian, Al-Dahle, Letman, Mathur, Schelten, Yang, Fan, Goyal, Hartshorn, Yang, Mitra, Sravankumar, Korenev, Hinsvark, Rao, Zhang et~al.}]{LLAMA3}
Abhimanyu Dubey, others Abhinav~Jauhri, Abhinav Pandey, Abhishek Kadian, Ahmad Al-Dahle, Aiesha Letman, Akhil Mathur, Alan Schelten, Amy Yang, Angela Fan, Anirudh Goyal, Anthony Hartshorn, Aobo Yang, Archi Mitra, Archie Sravankumar, Artem Korenev, Arthur Hinsvark, Arun Rao, Aston Zhang, et~al. 2024.
\newblock \href {https://arxiv.org/abs/2407.21783} {The llama 3 herd of models}.
\newblock \emph{Preprint}, arXiv:2407.21783.

\bibitem[{Fang et~al.(2024)Fang, Bai, Ni, Yang, Chen, and Xu}]{fang-etal-2024-enhancing}
Feiteng Fang, Yuelin Bai, Shiwen Ni, Min Yang, Xiaojun Chen, and Ruifeng Xu. 2024.
\newblock \href {https://doi.org/10.18653/v1/2024.acl-long.540} {Enhancing noise robustness of retrieval-augmented language models with adaptive adversarial training}.
\newblock In \emph{Proceedings of the 62nd Annual Meeting of the Association for Computational Linguistics (Volume 1: Long Papers)}, pages 10028--10039, Bangkok, Thailand. Association for Computational Linguistics.

\bibitem[{Glass et~al.(2022)Glass, Rossiello, Chowdhury, Naik, Cai, and Gliozzo}]{glass-etal-2022-re2g}
Michael Glass, Gaetano Rossiello, Md~Faisal~Mahbub Chowdhury, Ankita Naik, Pengshan Cai, and Alfio Gliozzo. 2022.
\newblock \href {https://doi.org/10.18653/v1/2022.naacl-main.194} {{R}e2{G}: Retrieve, rerank, generate}.
\newblock In \emph{Proceedings of the 2022 Conference of the North American Chapter of the Association for Computational Linguistics: Human Language Technologies}, pages 2701--2715, Seattle, United States. Association for Computational Linguistics.

\bibitem[{Guu et~al.(2020)Guu, Lee, Tung, Pasupat, and Chang}]{guu2020realm}
Kelvin Guu, Kenton Lee, Zora Tung, Panupong Pasupat, and Ming-Wei Chang. 2020.
\newblock \href {https://arxiv.org/abs/2002.08909} {Realm: Retrieval-augmented language model pre-training}.
\newblock \emph{Preprint}, arXiv:2002.08909.

\bibitem[{He et~al.(2024)He, Pan, Dong, Song, LiuYiBo, Qianguosun, Liang, Wang, Zhang, and Zhang}]{he-etal-2024-never}
Junqing He, Kunhao Pan, Xiaoqun Dong, Zhuoyang Song, LiuYiBo LiuYiBo, Qianguosun Qianguosun, Yuxin Liang, Hao Wang, Enming Zhang, and Jiaxing Zhang. 2024.
\newblock \href {https://doi.org/10.18653/v1/2024.acl-long.736} {Never lost in the middle: Mastering long-context question answering with position-agnostic decompositional training}.
\newblock In \emph{Proceedings of the 62nd Annual Meeting of the Association for Computational Linguistics (Volume 1: Long Papers)}, pages 13628--13642, Bangkok, Thailand. Association for Computational Linguistics.

\bibitem[{Hsieh et~al.(2024{\natexlab{a}})Hsieh, Sun, Kriman, Acharya, Rekesh, Jia, Zhang, and Ginsburg}]{hsieh2024ruler}
Cheng-Ping Hsieh, Simeng Sun, Samuel Kriman, Shantanu Acharya, Dima Rekesh, Fei Jia, Yang Zhang, and Boris Ginsburg. 2024{\natexlab{a}}.
\newblock Ruler: What's the real context size of your long-context language models?
\newblock \emph{arXiv preprint arXiv:2404.06654}.

\bibitem[{Hsieh et~al.(2024{\natexlab{b}})Hsieh, Chuang, Li, Wang, Le, Kumar, Glass, Ratner, Lee, Krishna, and Pfister}]{hsieh-etal-2024-found}
Cheng-Yu Hsieh, Yung-Sung Chuang, Chun-Liang Li, Zifeng Wang, Long Le, Abhishek Kumar, James Glass, Alexander Ratner, Chen-Yu Lee, Ranjay Krishna, and Tomas Pfister. 2024{\natexlab{b}}.
\newblock \href {https://doi.org/10.18653/v1/2024.findings-acl.890} {Found in the middle: Calibrating positional attention bias improves long context utilization}.
\newblock In \emph{Findings of the Association for Computational Linguistics ACL 2024}, pages 14982--14995, Bangkok, Thailand and virtual meeting. Association for Computational Linguistics.

\bibitem[{Huang et~al.(2023{\natexlab{a}})Huang, Yu, Ma, Zhong, Feng, Wang, Chen, Peng, Feng, Qin, and Liu}]{huang2023surveyhallucinationlargelanguage}
Lei Huang, Weijiang Yu, Weitao Ma, Weihong Zhong, Zhangyin Feng, Haotian Wang, Qianglong Chen, Weihua Peng, Xiaocheng Feng, Bing Qin, and Ting Liu. 2023{\natexlab{a}}.
\newblock \href {https://arxiv.org/abs/2311.05232} {A survey on hallucination in large language models: Principles, taxonomy, challenges, and open questions}.
\newblock \emph{Preprint}, arXiv:2311.05232.

\bibitem[{Huang et~al.(2023{\natexlab{b}})Huang, Yu, Ma, Zhong, Feng, Wang, Chen, Peng, Feng, Qin, and Liu}]{huang2023survey}
Lei Huang, Weijiang Yu, Weitao Ma, Weihong Zhong, Zhangyin Feng, Haotian Wang, Qianglong Chen, Weihua Peng, Xiaocheng Feng, Bing Qin, and Ting Liu. 2023{\natexlab{b}}.
\newblock \href {https://arxiv.org/abs/2311.05232} {A survey on hallucination in large language models: Principles, taxonomy, challenges, and open questions}.
\newblock \emph{Preprint}, arXiv:2311.05232.

\bibitem[{Izacard et~al.(2021)Izacard, Caron, Hosseini, Riedel, Bojanowski, Joulin, and Grave}]{izacard2021contriever}
Gautier Izacard, Mathilde Caron, Lucas Hosseini, Sebastian Riedel, Piotr Bojanowski, Armand Joulin, and Edouard Grave. 2021.
\newblock \href {https://doi.org/10.48550/ARXIV.2112.09118} {Unsupervised dense information retrieval with contrastive learning}.

\bibitem[{Izacard and Grave(2020)}]{izacard2020leveraging}
Gautier Izacard and Edouard Grave. 2020.
\newblock \href {https://arxiv.org/abs/2007.0128} {Leveraging passage retrieval with generative models for open domain question answering}.
\newblock \emph{arXiv preprint}.

\bibitem[{Izacard et~al.(2022)Izacard, Lewis, Lomeli, Hosseini, Petroni, Schick, Dwivedi-Yu, Joulin, Riedel, and Grave}]{izacard2022atlasfewshotlearningretrieval}
Gautier Izacard, Patrick Lewis, Maria Lomeli, Lucas Hosseini, Fabio Petroni, Timo Schick, Jane Dwivedi-Yu, Armand Joulin, Sebastian Riedel, and Edouard Grave. 2022.
\newblock \href {https://arxiv.org/abs/2208.03299} {Atlas: Few-shot learning with retrieval augmented language models}.
\newblock \emph{Preprint}, arXiv:2208.03299.

\bibitem[{Jin et~al.(2024)Jin, Zhang, Jiang, Liu, Liu, Liu, and Jin}]{jin2024ragcacheefficientknowledgecaching}
Chao Jin, Zili Zhang, Xuanlin Jiang, Fangyue Liu, Xin Liu, Xuanzhe Liu, and Xin Jin. 2024.
\newblock \href {https://arxiv.org/abs/2404.12457} {Ragcache: Efficient knowledge caching for retrieval-augmented generation}.
\newblock \emph{Preprint}, arXiv:2404.12457.

\bibitem[{Joshi et~al.(2017)Joshi, Choi, Weld, and Zettlemoyer}]{joshi-etal-2017-triviaqa}
Mandar Joshi, Eunsol Choi, Daniel Weld, and Luke Zettlemoyer. 2017.
\newblock \href {https://doi.org/10.18653/v1/P17-1147} {{T}rivia{QA}: A large scale distantly supervised challenge dataset for reading comprehension}.
\newblock In \emph{Proceedings of the 55th Annual Meeting of the Association for Computational Linguistics (Volume 1: Long Papers)}, pages 1601--1611, Vancouver, Canada. Association for Computational Linguistics.

\bibitem[{Kamradt(2023)}]{niah}
Gregory Kamradt. 2023.
\newblock \href {https://github.com/gkamradt/LLMTest_NeedleInAHaystack/tree/main} {Needle in a haystack - pressure testing llms.}

\bibitem[{Karpukhin et~al.(2020)Karpukhin, Oguz, Min, Lewis, Wu, Edunov, Chen, and Yih}]{karpukhin-etal-2020-dense}
Vladimir Karpukhin, Barlas Oguz, Sewon Min, Patrick Lewis, Ledell Wu, Sergey Edunov, Danqi Chen, and Wen-tau Yih. 2020.
\newblock \href {https://doi.org/10.18653/v1/2020.emnlp-main.550} {Dense passage retrieval for open-domain question answering}.
\newblock In \emph{Proceedings of the 2020 Conference on Empirical Methods in Natural Language Processing (EMNLP)}, pages 6769--6781, Online. Association for Computational Linguistics.

\bibitem[{Kwiatkowski et~al.(2019)Kwiatkowski, Palomaki, Redfield, Collins, Parikh, Alberti, Epstein, Polosukhin, Devlin, Lee, Toutanova, Jones, Kelcey, Chang, Dai, Uszkoreit, Le, and Petrov}]{kwiatkowski-etal-2019-natural}
Tom Kwiatkowski, Jennimaria Palomaki, Olivia Redfield, Michael Collins, Ankur Parikh, Chris Alberti, Danielle Epstein, Illia Polosukhin, Jacob Devlin, Kenton Lee, Kristina Toutanova, Llion Jones, Matthew Kelcey, Ming-Wei Chang, Andrew~M. Dai, Jakob Uszkoreit, Quoc Le, and Slav Petrov. 2019.
\newblock \href {https://doi.org/10.1162/tacl_a_00276} {Natural questions: A benchmark for question answering research}.
\newblock \emph{Transactions of the Association for Computational Linguistics}, 7:452--466.

\bibitem[{Lewis et~al.(2021)Lewis, Perez, Piktus, Petroni, Karpukhin, Goyal, Küttler, Lewis, tau Yih, Rocktäschel, Riedel, and Kiela}]{lewis2021retrievalaugmented}
Patrick Lewis, Ethan Perez, Aleksandra Piktus, Fabio Petroni, Vladimir Karpukhin, Naman Goyal, Heinrich Küttler, Mike Lewis, Wen tau Yih, Tim Rocktäschel, Sebastian Riedel, and Douwe Kiela. 2021.
\newblock \href {https://arxiv.org/abs/2005.11401} {Retrieval-augmented generation for knowledge-intensive nlp tasks}.
\newblock \emph{Preprint}, arXiv:2005.11401.

\bibitem[{Liu et~al.(2023)Liu, Lin, Hewitt, Paranjape, Bevilacqua, Petroni, and Liang}]{Liu2023LostIT}
Nelson~F. Liu, Kevin Lin, John Hewitt, Ashwin Paranjape, Michele Bevilacqua, Fabio Petroni, and Percy Liang. 2023.
\newblock \href {https://api.semanticscholar.org/CorpusID:259360665} {Lost in the middle: How language models use long contexts}.
\newblock \emph{Transactions of the Association for Computational Linguistics}, 12:157--173.

\bibitem[{Lu et~al.(2024)Lu, Wang, Rong, Chen, and Tang}]{lu2024turboragacceleratingretrievalaugmentedgeneration}
Songshuo Lu, Hua Wang, Yutian Rong, Zhi Chen, and Yaohua Tang. 2024.
\newblock \href {https://arxiv.org/abs/2410.07590} {Turborag: Accelerating retrieval-augmented generation with precomputed kv caches for chunked text}.
\newblock \emph{Preprint}, arXiv:2410.07590.

\bibitem[{Ma et~al.(2024)Ma, Wang, Yang, Wei, and Lin}]{10.1145/3626772.3657951}
Xueguang Ma, Liang Wang, Nan Yang, Furu Wei, and Jimmy Lin. 2024.
\newblock \href {https://doi.org/10.1145/3626772.3657951} {Fine-tuning llama for multi-stage text retrieval}.
\newblock In \emph{Proceedings of the 47th International ACM SIGIR Conference on Research and Development in Information Retrieval}, SIGIR '24, page 2421–2425, New York, NY, USA. Association for Computing Machinery.

\bibitem[{Mallen et~al.(2023)Mallen, Asai, Zhong, Das, Khashabi, and Hajishirzi}]{mallen2023trust}
Alex Mallen, Akari Asai, Victor Zhong, Rajarshi Das, Daniel Khashabi, and Hannaneh Hajishirzi. 2023.
\newblock \href {https://arxiv.org/abs/2212.10511} {When not to trust language models: Investigating effectiveness of parametric and non-parametric memories}.
\newblock \emph{Preprint}, arXiv:2212.10511.

\bibitem[{Menick et~al.(2022)Menick, Trebacz, Mikulik, Aslanides, Song, Chadwick, Glaese, Young, Campbell-Gillingham, Irving, and McAleese}]{menick2022teachinglanguagemodelssupport}
Jacob Menick, Maja Trebacz, Vladimir Mikulik, John Aslanides, Francis Song, Martin Chadwick, Mia Glaese, Susannah Young, Lucy Campbell-Gillingham, Geoffrey Irving, and Nat McAleese. 2022.
\newblock \href {https://arxiv.org/abs/2203.11147} {Teaching language models to support answers with verified quotes}.
\newblock \emph{Preprint}, arXiv:2203.11147.

\bibitem[{Nogueira et~al.(2020)Nogueira, Jiang, Pradeep, and Lin}]{nogueira-etal-2020-document}
Rodrigo Nogueira, Zhiying Jiang, Ronak Pradeep, and Jimmy Lin. 2020.
\newblock \href {https://doi.org/10.18653/v1/2020.findings-emnlp.63} {Document ranking with a pretrained sequence-to-sequence model}.
\newblock In \emph{Findings of the Association for Computational Linguistics: EMNLP 2020}, pages 708--718, Online. Association for Computational Linguistics.

\bibitem[{Oh and Thorne(2023)}]{oh-thorne-2023-detrimental}
Philhoon Oh and James Thorne. 2023.
\newblock \href {https://doi.org/10.18653/v1/2023.findings-emnlp.776} {Detrimental contexts in open-domain question answering}.
\newblock In \emph{Findings of the Association for Computational Linguistics: EMNLP 2023}, pages 11589--11605, Singapore. Association for Computational Linguistics.

\bibitem[{OpenAI(2024)}]{GPT-4o}
OpenAI. 2024.
\newblock \href {https://openai.com/index/hello-gpt-4o/} {Openai. 2024. gpt-4o}.

\bibitem[{Paszke et~al.(2017)Paszke, Gross, Chintala, Chanan, Yang, DeVito, Lin, Desmaison, Antiga, and Lerer}]{paszke2017automatic}
Adam Paszke, Sam Gross, Soumith Chintala, Gregory Chanan, Edward Yang, Zachary DeVito, Zeming Lin, Alban Desmaison, Luca Antiga, and Adam Lerer. 2017.
\newblock Automatic differentiation in pytorch.

\bibitem[{Petroni et~al.(2021)Petroni, Piktus, Fan, Lewis, Yazdani, De~Cao, Thorne, Jernite, Karpukhin, Maillard, Plachouras, Rockt{\"a}schel, and Riedel}]{petroni-etal-2021-kilt}
Fabio Petroni, Aleksandra Piktus, Angela Fan, Patrick Lewis, Majid Yazdani, Nicola De~Cao, James Thorne, Yacine Jernite, Vladimir Karpukhin, Jean Maillard, Vassilis Plachouras, Tim Rockt{\"a}schel, and Sebastian Riedel. 2021.
\newblock \href {https://doi.org/10.18653/v1/2021.naacl-main.200} {{KILT}: a benchmark for knowledge intensive language tasks}.
\newblock In \emph{Proceedings of the 2021 Conference of the North American Chapter of the Association for Computational Linguistics: Human Language Technologies}, pages 2523--2544, Online. Association for Computational Linguistics.

\bibitem[{Rudin et~al.(2021)Rudin, Chen, Chen, Huang, Semenova, and Zhong}]{rudin2021interpretable}
Cynthia Rudin, Chaofan Chen, Zhi Chen, Haiyang Huang, Lesia Semenova, and Chudi Zhong. 2021.
\newblock \href {https://arxiv.org/abs/2103.11251} {Interpretable machine learning: Fundamental principles and 10 grand challenges}.
\newblock \emph{Preprint}, arXiv:2103.11251.

\bibitem[{Shi et~al.(2023)Shi, Chen, Misra, Scales, Dohan, Chi, Schärli, and Zhou}]{shi2023largelanguagemodelseasily}
Freda Shi, Xinyun Chen, Kanishka Misra, Nathan Scales, David Dohan, Ed~Chi, Nathanael Schärli, and Denny Zhou. 2023.
\newblock \href {https://arxiv.org/abs/2302.00093} {Large language models can be easily distracted by irrelevant context}.
\newblock \emph{Preprint}, arXiv:2302.00093.

\bibitem[{Shi et~al.(2024{\natexlab{a}})Shi, Han, Lewis, Tsvetkov, Zettlemoyer, and Yih}]{shi-etal-2024-trusting}
Weijia Shi, Xiaochuang Han, Mike Lewis, Yulia Tsvetkov, Luke Zettlemoyer, and Wen-tau Yih. 2024{\natexlab{a}}.
\newblock \href {https://doi.org/10.18653/v1/2024.naacl-short.69} {Trusting your evidence: Hallucinate less with context-aware decoding}.
\newblock In \emph{Proceedings of the 2024 Conference of the North American Chapter of the Association for Computational Linguistics: Human Language Technologies (Volume 2: Short Papers)}, pages 783--791, Mexico City, Mexico. Association for Computational Linguistics.

\bibitem[{Shi et~al.(2024{\natexlab{b}})Shi, Min, Yasunaga, Seo, James, Lewis, Zettlemoyer, and Yih}]{shi-etal-2024-replug}
Weijia Shi, Sewon Min, Michihiro Yasunaga, Minjoon Seo, Richard James, Mike Lewis, Luke Zettlemoyer, and Wen-tau Yih. 2024{\natexlab{b}}.
\newblock \href {https://doi.org/10.18653/v1/2024.naacl-long.463} {{REPLUG}: Retrieval-augmented black-box language models}.
\newblock In \emph{Proceedings of the 2024 Conference of the North American Chapter of the Association for Computational Linguistics: Human Language Technologies (Volume 1: Long Papers)}, pages 8371--8384, Mexico City, Mexico. Association for Computational Linguistics.

\bibitem[{Sun et~al.(2024)Sun, Dong, Zhu, Huang, Wang, Ma, Zhang, Wang, and Wei}]{sun2024cacheoncedecoderdecoderarchitectures}
Yutao Sun, Li~Dong, Yi~Zhu, Shaohan Huang, Wenhui Wang, Shuming Ma, Quanlu Zhang, Jianyong Wang, and Furu Wei. 2024.
\newblock \href {https://arxiv.org/abs/2405.05254} {You only cache once: Decoder-decoder architectures for language models}.
\newblock \emph{Preprint}, arXiv:2405.05254.

\bibitem[{Team et~al.(2024)Team, Georgiev, Lei, Burnell, Bai, Gulati, Tanzer, Vincent, Pan, Wang, Mariooryad, Ding et~al.}]{geminiteam2024gemini15unlockingmultimodal}
Gemini Team, Petko Georgiev, Ving~Ian Lei, Ryan Burnell, Libin Bai, Anmol Gulati, Garrett Tanzer, Damien Vincent, Zhufeng Pan, Shibo Wang, Soroosh Mariooryad, Yifan Ding, et~al. 2024.
\newblock \href {https://arxiv.org/abs/2403.05530} {Gemini 1.5: Unlocking multimodal understanding across millions of tokens of context}.
\newblock \emph{Preprint}, arXiv:2403.05530.

\bibitem[{Thoppilan et~al.(2022)Thoppilan, Freitas, Hall, Shazeer, Kulshreshtha, Cheng, Jin, Bos, Baker, Du, Li, Lee, Zheng, Ghafouri, Menegali, Huang, Krikun, Lepikhin, Qin, Chen, Xu, Chen, Roberts, Bosma, Zhao, Zhou, Chang, Krivokon, Rusch, Pickett, Srinivasan, Man, Meier-Hellstern, Morris, Doshi, Santos, Duke, Soraker, Zevenbergen, Prabhakaran, Diaz, Hutchinson, Olson, Molina, Hoffman-John, Lee, Aroyo, Rajakumar, Butryna, Lamm, Kuzmina, Fenton, Cohen, Bernstein, Kurzweil, Aguera-Arcas, Cui, Croak, Chi, and Le}]{thoppilan2022lamdalanguagemodelsdialog}
Romal Thoppilan, Daniel~De Freitas, Jamie Hall, Noam Shazeer, Apoorv Kulshreshtha, Heng-Tze Cheng, Alicia Jin, Taylor Bos, Leslie Baker, Yu~Du, YaGuang Li, Hongrae Lee, Huaixiu~Steven Zheng, Amin Ghafouri, Marcelo Menegali, Yanping Huang, Maxim Krikun, Dmitry Lepikhin, James Qin, Dehao Chen, Yuanzhong Xu, Zhifeng Chen, Adam Roberts, Maarten Bosma, Vincent Zhao, Yanqi Zhou, Chung-Ching Chang, Igor Krivokon, Will Rusch, Marc Pickett, Pranesh Srinivasan, Laichee Man, Kathleen Meier-Hellstern, Meredith~Ringel Morris, Tulsee Doshi, Renelito~Delos Santos, Toju Duke, Johnny Soraker, Ben Zevenbergen, Vinodkumar Prabhakaran, Mark Diaz, Ben Hutchinson, Kristen Olson, Alejandra Molina, Erin Hoffman-John, Josh Lee, Lora Aroyo, Ravi Rajakumar, Alena Butryna, Matthew Lamm, Viktoriya Kuzmina, Joe Fenton, Aaron Cohen, Rachel Bernstein, Ray Kurzweil, Blaise Aguera-Arcas, Claire Cui, Marian Croak, Ed~Chi, and Quoc Le. 2022.
\newblock \href {https://arxiv.org/abs/2201.08239} {Lamda: Language models for dialog applications}.
\newblock \emph{Preprint}, arXiv:2201.08239.

\bibitem[{Thorne et~al.(2018)Thorne, Vlachos, Christodoulopoulos, and Mittal}]{thorne-etal-2018-fever}
James Thorne, Andreas Vlachos, Christos Christodoulopoulos, and Arpit Mittal. 2018.
\newblock \href {https://doi.org/10.18653/v1/N18-1074} {{FEVER}: a large-scale dataset for fact extraction and {VER}ification}.
\newblock In \emph{Proceedings of the 2018 Conference of the North {A}merican Chapter of the Association for Computational Linguistics: Human Language Technologies, Volume 1 (Long Papers)}, pages 809--819, New Orleans, Louisiana. Association for Computational Linguistics.

\bibitem[{Trivedi et~al.(2022)Trivedi, Balasubramanian, Khot, and Sabharwal}]{trivedi-etal-2022-musique}
Harsh Trivedi, Niranjan Balasubramanian, Tushar Khot, and Ashish Sabharwal. 2022.
\newblock \href {https://doi.org/10.1162/tacl_a_00475} {Musique: Multihop questions via single-hop question composition}.
\newblock \emph{Transactions of the Association for Computational Linguistics}, 10:539--554.

\bibitem[{Wolf et~al.(2020)Wolf, Debut, Sanh, Chaumond, Delangue, Moi, Cistac, Rault, Louf, Funtowicz, Davison, Shleifer, von Platen, Ma, Jernite, Plu, Xu, Scao, Gugger, Drame, Lhoest, and Rush}]{wolf2020huggingfacestransformersstateoftheartnatural}
Thomas Wolf, Lysandre Debut, Victor Sanh, Julien Chaumond, Clement Delangue, Anthony Moi, Pierric Cistac, Tim Rault, Rémi Louf, Morgan Funtowicz, Joe Davison, Sam Shleifer, Patrick von Platen, Clara Ma, Yacine Jernite, Julien Plu, Canwen Xu, Teven~Le Scao, Sylvain Gugger, Mariama Drame, Quentin Lhoest, and Alexander~M. Rush. 2020.
\newblock \href {https://arxiv.org/abs/1910.03771} {Huggingface's transformers: State-of-the-art natural language processing}.
\newblock \emph{Preprint}, arXiv:1910.03771.

\bibitem[{Xu et~al.(2023)Xu, Shi, and Choi}]{xu2023recompimprovingretrievalaugmentedlms}
Fangyuan Xu, Weijia Shi, and Eunsol Choi. 2023.
\newblock \href {https://arxiv.org/abs/2310.04408} {Recomp: Improving retrieval-augmented lms with compression and selective augmentation}.
\newblock \emph{Preprint}, arXiv:2310.04408.

\bibitem[{Yang et~al.(2018)Yang, Qi, Zhang, Bengio, Cohen, Salakhutdinov, and Manning}]{yang2018hotpotqa}
Zhilin Yang, Peng Qi, Saizheng Zhang, Yoshua Bengio, William~W. Cohen, Ruslan Salakhutdinov, and Christopher~D. Manning. 2018.
\newblock {HotpotQA}: A dataset for diverse, explainable multi-hop question answering.
\newblock In \emph{Conference on Empirical Methods in Natural Language Processing ({EMNLP})}.

\bibitem[{Yen et~al.(2024)Yen, Gao, and Chen}]{yen2024longcontextlanguagemodelingparallel}
Howard Yen, Tianyu Gao, and Danqi Chen. 2024.
\newblock \href {https://arxiv.org/abs/2402.16617} {Long-context language modeling with parallel context encoding}.
\newblock \emph{Preprint}, arXiv:2402.16617.

\bibitem[{Yoran et~al.(2024{\natexlab{a}})Yoran, Wolfson, Ram, and Berant}]{yoran2024making}
Ori Yoran, Tomer Wolfson, Ori Ram, and Jonathan Berant. 2024{\natexlab{a}}.
\newblock \href {https://openreview.net/forum?id=ZS4m74kZpH} {Making retrieval-augmented language models robust to irrelevant context}.
\newblock In \emph{The Twelfth International Conference on Learning Representations}.

\bibitem[{Yoran et~al.(2024{\natexlab{b}})Yoran, Wolfson, Ram, and Berant}]{yoran2024makingretrievalaugmentedlanguagemodels}
Ori Yoran, Tomer Wolfson, Ori Ram, and Jonathan Berant. 2024{\natexlab{b}}.
\newblock \href {https://arxiv.org/abs/2310.01558} {Making retrieval-augmented language models robust to irrelevant context}.
\newblock \emph{Preprint}, arXiv:2310.01558.

\bibitem[{Yu et~al.(2024{\natexlab{a}})Yu, Jiang, Luo, Wu, Lin, Li, Yang, Huang, and Qiu}]{yu2024mitigatepositionbiaslarge}
Yijiong Yu, Huiqiang Jiang, Xufang Luo, Qianhui Wu, Chin-Yew Lin, Dongsheng Li, Yuqing Yang, Yongfeng Huang, and Lili Qiu. 2024{\natexlab{a}}.
\newblock \href {https://arxiv.org/abs/2406.02536} {Mitigate position bias in large language models via scaling a single dimension}.
\newblock \emph{Preprint}, arXiv:2406.02536.

\bibitem[{Yu et~al.(2024{\natexlab{b}})Yu, Ping, Liu, Wang, You, Zhang, Shoeybi, and Catanzaro}]{yu2024rankragunifyingcontextranking}
Yue Yu, Wei Ping, Zihan Liu, Boxin Wang, Jiaxuan You, Chao Zhang, Mohammad Shoeybi, and Bryan Catanzaro. 2024{\natexlab{b}}.
\newblock \href {https://openreview.net/forum?id=S1fc92uemC} {Rank{RAG}: Unifying context ranking with retrieval-augmented generation in {LLM}s}.
\newblock In \emph{The Thirty-eighth Annual Conference on Neural Information Processing Systems}.

\bibitem[{Zhuang et~al.(2023)Zhuang, Qin, Jagerman, Hui, Ma, Lu, Ni, Wang, and Bendersky}]{10.1145/3539618.3592047}
Honglei Zhuang, Zhen Qin, Rolf Jagerman, Kai Hui, Ji~Ma, Jing Lu, Jianmo Ni, Xuanhui Wang, and Michael Bendersky. 2023.
\newblock \href {https://doi.org/10.1145/3539618.3592047} {Rankt5: Fine-tuning t5 for text ranking with ranking losses}.
\newblock In \emph{Proceedings of the 46th International ACM SIGIR Conference on Research and Development in Information Retrieval}, SIGIR '23, page 2308–2313, New York, NY, USA. Association for Computing Machinery.

\end{thebibliography}

\clearpage
\onecolumn
\appendix

\section{Appendix}
\label{sec:appendix}
\subsection{Details for KV-Fusion Implementation}\label{appendix:pseudo}
This section describes the pseudocode for KV-Fusion and Python implementation of the \reshape\ function.

\subsubsection{Algorithm for KV-Fusion}
This section further elaborates on the KV-Fusion algorithm, which can be implemented using standard language modeling. For clarification, we provide the pseudocode with a single-instance example. As explained in Section~\ref{sec:method}, KV-Fusion is built upon two decoders. First, $\df$ processes a set of input passages retrieved by retrievers, ${C} = \{c_1, c_2, \dots, c_N\}$, and generates Key-Value (KV) caches in parallel. These KV caches are reshaped by the \reshape\ function into the form $\{K^{l}, V^{l}\}_{l=1}^L$ to prefill the cache in $\dt$. Next, $\dt$ processes target tokens, $t = \{t_1, t_2, \dots, t_m\}$, along with their positional information, $p = \{p_{n+1}, p_{n+2}, \dots, p_{n+m}\}$. Specifically, the target tokens consist of two parts: the query part, which includes instructions, $q = \{t_1, t_2, \dots, t_k\}$, and the answer part, $y = \{t_{k+1}, t_{k+2}, \dots, t_m\}$. Along with the prefilled KV cache, we train $\dt$ by prompting it with $q$ and using the standard language model loss to generate $y$. For implementation, we use \texttt{huggingface transformers}\cite{wolf2020huggingfacestransformersstateoftheartnatural} and \texttt{PyTorch}\cite{paszke2017automatic} libraries.

\begin{figure}[ht!]
\centering

\begin{algorithm}[H]
\caption{Key Value Fusion(KV Fusion)}
\label{alg:aggregate}
 \textbf{Input:} $\dt$, $\df$, Training Data $\mathcal{C^{\mathtt{train}}} = \{C_1, C_2, \dots, C_L\}$ where ${C_i} = \{c^{i}_1, c^{i}_2, \dots, c^{i}_N\}$, Corresponding query tokens $q_{i} = \{t^{i}_1, t^{i}_2, \dots, t^{i}_k\}$, Corresponding answer tokens $y_{i} = \{t^{i}_{k+1}, t^{i}_{k+2}, \dots, t^{i}_m\}$ \\
\vspace{-0.15in}

\begin{algorithmic}[1]
\State Initialize $\dt$, $\df$ and freeze $\df$
\For{$i=1,2,...,L$} \do
    
    \State{\tt \#  Extract KV-caches in parallel}
    \State  ${KV}_{\mathtt{cache}}=\df(C_i)$
    \State 
    \State{\tt \#  Reshape KV-caches}  
    \State $\{K^{l}, V^{l}\}_{l=1}^L=\reshape({KV}_{\mathtt{cache}})$
    \State
    \State{\tt \#  Compute loss and Optimize $\dt$} 
    \State $Loss = \mathcal{LM}_{\mathtt{loss}}(y_{i}, \dt(q_{i}; \{K^{l}, V^{l}\}_{l=1}^L))$
    \State Update parameters of $\dt$ with respect to $Loss$ via gradient descent

\EndFor
 
\end{algorithmic}

\end{algorithm}
\end{figure}

\subsubsection{$\reshape$ Implementation}
To train $\dt$ seamlessly with \texttt{huggingface transformers}\cite{wolf2020huggingfacestransformersstateoftheartnatural} and \texttt{PyTorch}\cite{paszke2017automatic}, extracted KV-cache need to be reshaped to prefill the caches in $\dt$. To this end, we implement \reshape\ function down below, which can also process batch of instances.   

\begin{scriptsize}
\begin{center}
\begin{BVerbatim}[frame=single]
def reshape_key_value_batches(cur_past_key_values, n_psgs):
    """
    Reshape key-value pairs in batches
    """
    new_key_cache = []

    # Iterate through each key-value pair
    for k, v in cur_past_key_values:
        # Split keys and values into splits (split by instance)
        k_splits, v_splits = torch.split(k, n_psgs, dim=0), torch.split(v, n_psgs, dim=0)

        # Reshape and concatenate splits (reshape by instance)
        k_re = torch.cat([torch.cat(torch.split(k_val, 1, dim=0), dim=2) for k_val in k_splits], dim=0)
        v_re = torch.cat([torch.cat(torch.split(v_val, 1, dim=0), dim=2) for v_val in v_splits], dim=0)

        # Append processed key-value pair
        new_key_cache.append((k_re, v_re))

    return tuple(new_key_cache)
\end{BVerbatim}
\end{center}
\end{scriptsize}

\clearpage

\subsection{Dataset Construction}\label{appendix:data_cons}
\subsubsection{Prompt Template}\label{appendix:ext_gold}
\begin{mybox}{Prompt Template for Verifying Gold Passages and Supporting Evidence}
Your task is to find Evidence from a given Document based on a Question and its corresponding Answer. Specifically, the Document contains the Answer for the given Question. Your job is to extract the Evidence from the document.\\
\\
Here are the Question, Document, and Answer.\\
\\
Question:\\
\{QUESTION\}\\
\\
Document:\\
\{PASSAGE\}\\
\\
Answer:\\
\{ANSWER\}\\
\\
Here is how the evidence should be presented:\\
\\
* Evidence \\
- The Evidence should only consist of sentences or paraphrases taken from the given Document. \\
- The Evidence should retain the same format as in the given Document.\\
- The Evidence should inlcude enough information to derive the given Answer from the given Question.\\
- If the provided Document does not contain sufficient information, generate NONE.\\
\\
* Format\\
- DO NOT WRITE ANY GREETING MESSAGES, just write the evidence only.\\
- In front of the evidence, append the word ``Evidence:".\\
- Write [END] after you are done.\\
- Here is the Example Format:\\
``\\
Evidence: evidence sentences [END]\\
``\\
- Do not include `` in the response.\\
\\
Data Generation:
\end{mybox}

\vspace{2.5mm}

\subsubsection{Dataset Statistics}\label{appendix:dataset}
\begin{table}[htbp]
\centering
\begin{tabular}{|c|c|c|c|}
\hline
\textbf{Dataset} & \textbf{Training} & \textbf{Dev} & \textbf{Test} \\ \hline
\textbf{NQ}      & 47,633            & 3,036        & 3,610         \\ \hline
\textbf{TQA}     & 34,648            & 4,288        & 1,768         \\ \hline
\textbf{POPQA}     & 6,833             & 1,190        & 1,267         \\ \hline
\end{tabular}
\caption{Dataset Statistics for NQ, TQA, and PQA}
\label{tab:dataset_stats}
\end{table}
NQ and TriviaQA are filtered versions provided by \citealt{karpukhin-etal-2020-dense} under the CC BY-NC 4.0. We also use POPQA, available from the HuggingFace datasets\footnote{https://huggingface.co/datasets/akariasai/PopQA} under MIT License. These datasets are English open domain question answering datasets based on the December 2018 Wikipedia snapshot, preprocessed by \citealt{karpukhin-etal-2020-dense}.

\begin{itemize}
\item NQ dataset: Following the procedure outlined in Section~\ref{experiments:dataset}, we obtained 47,633, 3,036, and 3,610 instances for the training, dev, and test sets, respectively.
\item TQA dataset: To manage the GPT-4o API budget, the original TQA dev set was split in a 2:1 ratio, resulting in newly defined dev and test sets. This produced 34,648, 4,288, and 1,768 instances for the training, dev, and test sets, respectively.
\item POPQA dataset: The original dataset did not include pre-defined training or dev sets. We split the data in an 8:1:1 ratio. After processing with the GPT-4 API, this resulted in 6,833, 1,190, and 1,267 instances for the training, dev, and test sets, respectively
\end{itemize}

\vspace{3mm}

\subsection{Baseline Model Template and Example}\label{appendix:baseline_format}
\begin{mybox}{Baseline Template}
Title: \{TITLE\} Context: \{TEXT\}\\
=====\\
Title: \{TITLE\} Context: \{TEXT\}\\
=====\\
.\\
.\\
.\\
=====\\
Title: \{TITLE\} Context: \{TEXT\}\\
=====\\
Strictly based on listed documents (titles and contexts) above, answer the given question clearly and concisely in a single sentence. If none of the documents provide a valid answer, respond with ``Unanswerable''. Question: \{QUESTION\}? ANSWER:
\end{mybox}

\begin{mybox}{Baseline Example}
Title: Nobel Prize in Physics Context: Nobel Prize in Physics The Nobel Prize in Physics ()  is a yearly award given by the Royal Swedish Academy of Sciences for...\\
=====\\
Title: Nobel Prize Context: His son, George Paget Thomson, received the same prize in 1937 for showing that they also have the properties of waves...\\
=====\\
.\\
.\\
.\\
=====\\
Title: Nobel Prize Context: Wilhelm Röntgen's discovery of X-rays and Philipp Lenard's work on cathode rays.  The Academy of Sciences selected Röntgen...\\
=====\\
Strictly based on listed documents (titles and contexts) above, answer the given question clearly and concisely in a single sentence. If none of the documents provide a valid answer, respond with ``Unanswerable''. Question: who got the first nobel prize in physics? ANSWER:
\end{mybox}

\clearpage

\subsection{KV-Model Input Format Template and Example}\label{appendix:format}
The following examples outline the input format template along with a concrete example for $\df$.
\begin{mybox}{Input Template for $\df$}
Title: \{TITLE\} Context: \{TEXT\}\\
=====
\end{mybox}

\begin{mybox}{Input Example for $\df$}
Title: Does He Love You Context: Does He Love You "Does He Love You" is a song written by Sandy Knox and Billy Stritch, and recorded as a duet by American country music artists Reba McEntire and Linda Davis. It was released in August 1993 as the first single from Reba's album "Greatest Hits Volume Two". It is one of country music's several songs about a love triangle. "Does He Love You" was written in 1982 by Billy Stritch. He recorded it with a trio in which he performed at the time, because he wanted a song that could be sung by the other two members\\
=====
\end{mybox}

The following example outlines the input format template along with a concrete example for $\dt$.

\begin{mybox}{Input Template for $\dt$}
{\small{\texttt{$<$|question\_answering|$>$}}} Using the provided titles and contexts, answer the given question briefly and provide the supporting sentences as evidence. \\
Question: \{QUESTION\}? \\ 
Answer: \{ANSWER\} [RESULT] \\
Evidence: \{EVIDENCE\} [END]
\end{mybox}

\begin{mybox}{Input Example for $\dt$}
{\small{\texttt{$<$|question\_answering|$>$}}} Using the provided titles and contexts, answer the given question briefly and provide the supporting sentences as evidence. \\
Question: who sings does he love me with reba? \\ 
Answer: Linda Davis [RESULT] \\
Evidence: "Does He Love You" is a song written by Sandy Knox and Billy Stritch, and recorded as a duet by American country music artists Reba McEntire and Linda Davis. [END]
\end{mybox}

\clearpage

\subsection{Hyperparameters for training}\label{appendix:hyper}

\begin{table}[htbp]
\renewcommand{\arraystretch}{0.9}
\centering
\begin{tabular}{|c|c|}
\hline
\multicolumn{2}{|c|}{\textbf{Training Hyperparameters}} \\ \hline
flash attention 2   & True         \\ \hline
target token max length    & 192 \\ \hline
number of input contexts    & 20 \\ \hline
input token max length    & 192 \\ \hline
epochs    & 2 \\ \hline
batch size per gpu    & 2 \\ \hline
gradient accumulation & 8 \\ \hline
learning rate & 2e-5 \\ \hline
warmup ratio & 0.05 \\ \hline
scheduler & cosine \\ \hline
optimizer  & adamW  \\ \hline
\end{tabular}
\caption{Across all datasets, we utilize four A100 80 GB GPUs, with a batch size of 2 per device and a gradient accumulation of 8, consuming approximately 12 hours (48 GPU hours). The number of training passages is 20, consisting of one gold context and 19 negative contexts. Contexts are tokenized with a maximum length of 192 tokens using left padding; if a context exceeds this limit, tokens are truncated from the left. A random sample of 10,000 contexts from the NQ training set showed that 99.0\% of contexts fit within this token limit.(Avg: 145 tokens, Std: 14 tokens) The maximum learning rate is set to $2 \times 10^{-5}$, using a linear warmup and cosine decay. The warmup ratio is set to 5\%. The AdamW optimizer(paged\_adamw\_32bit) is used with $\beta_1=0.9$ and $\beta_2=0.999$.}
\vspace{-5mm}
\label{tab:hyper}
\end{table}

\subsection{KV-Llama3.1 and Llama3.1}\label{appendix:llama3.1}
\begin{figure*}[!ht]
    \centering

    \begin{subfigure}{0.3\textwidth}
        \centering
        \caption{\scriptsize NQ - Llama3.1}
        \includegraphics[width=\textwidth]{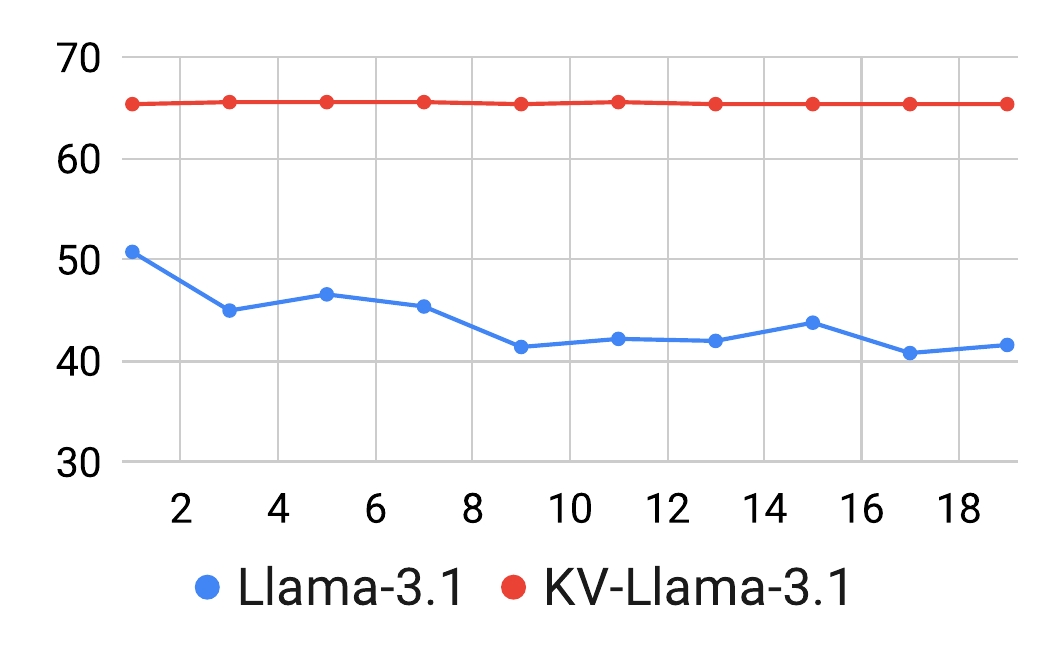}
    \end{subfigure}
    \hfill
    \begin{subfigure}{0.3\textwidth}
        \centering
        \caption{\scriptsize TQA - Llama3.1}
        \includegraphics[width=\textwidth]{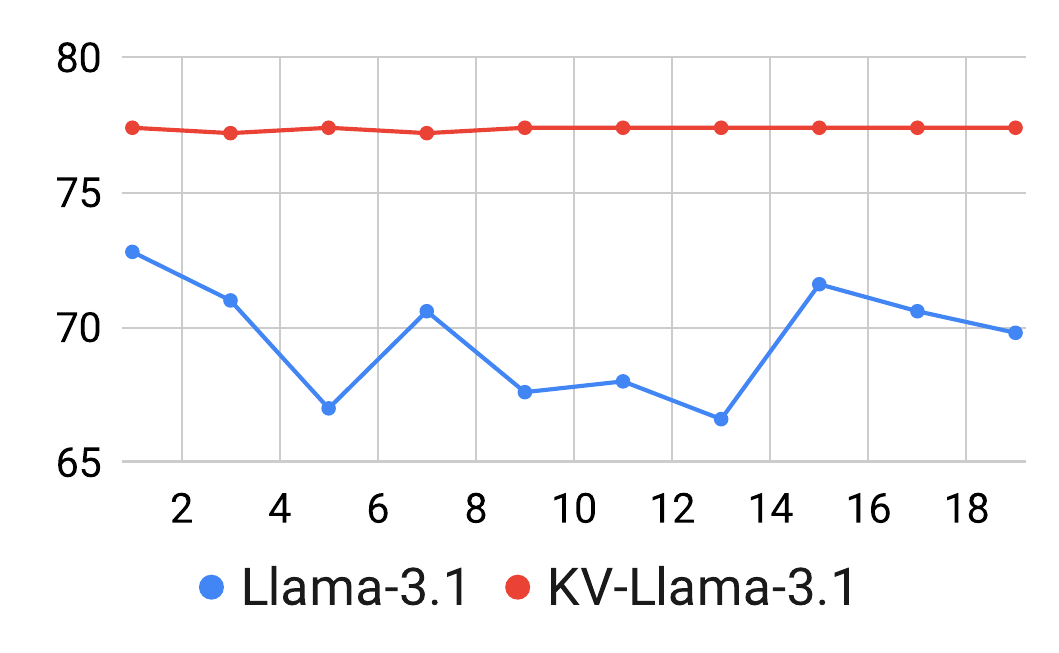}
    \end{subfigure}
    \hfill
    \begin{subfigure}{0.3\textwidth}
        \centering
        \caption{\scriptsize POPQA - Llama3.1}
        \includegraphics[width=\textwidth]{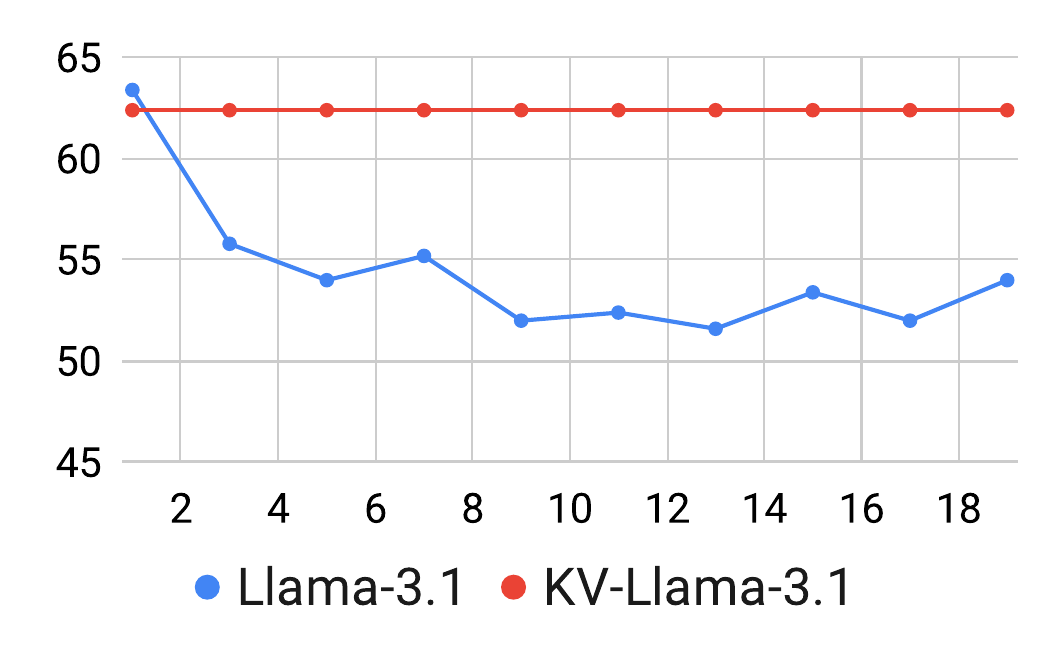}
    \end{subfigure}


    \caption{Comparison of EM Accuracy between \textcolor{red}{KV-Llama3.1} and \textcolor{blue}{Llama3.1} across different gold context positions. With varying gold context positions, KV-Llama3.1 illustrates consistent accuracies across datasets. However, Llama3.1 suffers from the `Lost in the middle' problem, which can be resolved by KV-Fusion models.}
    \vspace{-5mm}
    \label{fig:exp1-1-llama3.1}
\end{figure*}


\subsection{Position Agnostic reader evaluation on Contriever retrieved passages}\label{appendix:contriever}
\begin{table*}[!ht]
    \renewcommand{\arraystretch}{1.15}
    \centering
    \small
    \begin{tabularx}{\textwidth}{p{3.1cm} XXXXX XXXXX XXXXX}
        \toprule
        \textbf{Dataset}    & \multicolumn{4}{c}{\textbf{NQ}}   & \multicolumn{4}{c}{\textbf{TQA}}  & \multicolumn{4}{c}{\textbf{POPQA}}  \\
        \cmidrule(lr){2-5} \cmidrule(lr){6-9} \cmidrule(lr){10-13} 
        \textbf{Top-K} & 5 & 10 & 20 & 40 & 5 & 10 & 20 & 40 & 5 & 10 & 20 & 40    \\
        \midrule
        \llama  &\scriptsize{29.6}  &\scriptsize{32.0} &\scriptsize{36.6} &\scriptsize{36.1}  &\scriptsize{55.1} &\scriptsize{59.7} &\scriptsize{61.8} &\scriptsize{58.7}  &\scriptsize{37.5} &\scriptsize{39.1} &\scriptsize{37.7} &\scriptsize{36.9}\\
        \llamaone  &\scriptsize{38.8}  &\scriptsize{36.2} &\scriptsize{37.7} &\scriptsize{38.4}   &\scriptsize{58.4} &\scriptsize{57.6} &\scriptsize{59.4} &\scriptsize{60.7} &\scriptsize{38.8} &\scriptsize{38.8} &\scriptsize{38.4} &\scriptsize{38.8}\\
        \hdashline
        \replug &\scriptsize{33.4}  &\scriptsize{33.1} &\scriptsize{31.3} &\scriptsize{31.0}  &\scriptsize{54.2}  &\scriptsize{54.5} &\scriptsize{55.2} &\scriptsize{55.8} &\scriptsize{32.4}  &\scriptsize{28.0} &\scriptsize{25.8} &\scriptsize{23.9} \\
        \replugone  &\scriptsize{35.3}  &\scriptsize{34.9} &\scriptsize{33.4} &\scriptsize{32.7}  &\scriptsize{61.0}  &\scriptsize{60.9} &\scriptsize{60.1} &\scriptsize{60.4} &\scriptsize{39.9}  &\scriptsize{34.3} &\scriptsize{30.8} &\scriptsize{28.8} \\
        \PAMQA  &\textbf{\scriptsize{49.9}}  &\scriptsize{44.1} &\scriptsize{38.1} &\scriptsize{18.9}  &\scriptsize{64.4}  &\scriptsize{57.7} &\scriptsize{52.7} &\scriptsize{28.5} &\scriptsize{51.0} &\scriptsize{48.8} &\scriptsize{44.0} &\scriptsize{24.2} \\
        \kvllamaT  &\scriptsize{49.1}  &\scriptsize{50.6} &\textbf{\scriptsize{50.3}} &\textbf{\scriptsize{49.1}} &\scriptsize{65.1}  &\textbf{\scriptsize{67.1}} &\scriptsize{68.4} &\textbf{\scriptsize{68.3}} &\textbf{\scriptsize{53.9}} &\textbf{\scriptsize{56.6}} &\textbf{\scriptsize{54.5}} &\textbf{\scriptsize{51.9}} \\
        \kvllamaoneT  &\scriptsize{48.9}  &\textbf{\scriptsize{50.9}} &\textbf{\scriptsize{50.3}} &\scriptsize{48.9}  &\textbf{\scriptsize{65.7}}  &\scriptsize{66.3} &\textbf{\scriptsize{68.8}} &\scriptsize{67.8} &\scriptsize{53.0} &\scriptsize{54.8} &\scriptsize{54.1} &\scriptsize{51.3} \\
        \bottomrule
    \end{tabularx}
    \caption{Accuracy comparison with other position-invariant methods on contriever-retrieved passages. KV-Fusion models achieve the highest accuracies across datasets except for NQ top-5 case. Consistent with the results observed for DPR-retrieved passages in Table~\ref{tab:exp2}, KV-Fusion models show strong robustness to the inclusion of additional passages while other methods experience a decline in performance as more passages are added. Notably, the strong performance on POPQA datasets highlights Contriever's ability to excel on unseen datasets, which KV-Fusion models effectively leverage. This demonstrates that KV-Fusion models can achieve strong performance on different retrievers.}
    
    \vspace{-5mm}
    \label{tab:exp22}
\end{table*}

\end{document}